\documentclass[sigconf,authorversion]{acmart}
\copyrightyear{2021} 
\acmYear{2021} 
\setcopyright{rightsretained} 
\acmConference[ACM CHIL '21]{ACM Conference on Health, Inference, and Learning}{April 8--10, 2021}{Virtual Event, USA}
\acmBooktitle{ACM Conference on Health, Inference, and Learning (ACM CHIL '21), April 8--10, 2021, Virtual Event, USA}\acmDOI{10.1145/3450439.3451866}
\acmISBN{978-1-4503-8359-2/21/04}

\usepackage{subfigure}

\usepackage[makeroom]{cancel}
\usepackage{algorithm}
\usepackage{algorithmic}
\usepackage{multirow}
\usepackage{amsfonts}       % blackboard math symbols
\usepackage{hyperref}

\newcommand\independent{\protect\mathpalette{\protect\independenT}{\perp}}
\def\independenT#1#2{\mathrel{\rlap{$#1#2$}\mkern2mu{#1#2}}}
\newcommand\tf{\theta_f}
\newcommand\ttf{\tilde{\theta}_f}
\newcommand\tz{\tilde{z}_0}
\newcommand\htf{\hat{\theta}^i_f}
\newcommand\htz{\hat{z}^i_0}
\newcommand{\cmark}{\checkmark}%
\newcommand{\xmark}{\text{\sffamily X}}
\usepackage{array}
\newcommand{\PreserveBackslash}[1]{\let\temp=\\#1\let\\=\temp}
\newcolumntype{C}[1]{>{\PreserveBackslash\centering}p{#1}}
\newcolumntype{R}[1]{>{\PreserveBackslash\raggedleft}p{#1}}
\newcolumntype{L}[1]{>{\PreserveBackslash\raggedright}p{#1}}

\newcommand{\tabref}[1]{Table \ref{#1}}
\newcommand{\figref}[1]{Fig. \ref{#1}}
\newcommand{\algref}[1]{Algorithm \ref{#1}}

\begin{document}

\title{Generative ODE Modeling with Known Unknowns}


\author{Ori Linial}
\email{linial04@campus.technion.ac.il}
\affiliation{%
  \institution{Technion - Israel Institute of Technology}
  \city{Haifa}
  \country{Israel}
}

\author{Neta Ravid}
\email{neta.r@technion.ac.il}
\affiliation{%
  \institution{Technion - Israel Institute of Technology}
  \city{Haifa}
  \country{Israel}
}

\author{Danny Eytan}
\email{danny.eytan@technion.ac.il}
\affiliation{%
  \institution{Technion - Israel Institute of Technology}
  \institution{Rambam Health Care Campus}
  \city{Haifa}
  \country{Israel}
}

\author{Uri Shalit}
\email{urishalit@technion.ac.il}
\affiliation{%
  \institution{Technion - Israel Institute of Technology}
  \city{Haifa}
  \country{Israel}
}

\begin{abstract}
In several crucial applications, domain knowledge is encoded by a system of ordinary differential equations (ODE), often stemming from underlying physical and biological processes. A motivating example is intensive care unit patients: the dynamics of vital physiological functions, such as the cardiovascular system with its associated variables (heart rate, cardiac contractility and output and vascular resistance) can be approximately described by a known system of ODEs. Typically, some of the ODE variables are directly observed (heart rate and blood pressure for example) while some are unobserved (cardiac contractility, output and vascular resistance), and in addition many other variables are observed but not modeled by the ODE, for example body temperature. Importantly, the unobserved ODE variables are ``known-unknowns'': We know they exist and their functional dynamics, but cannot measure them directly, nor do we know the function tying them to all observed measurements. As is often the case in medicine, and specifically the cardiovascular system, estimating these known-unknowns is highly valuable and they serve as targets for therapeutic manipulations. 
Under this scenario we wish to learn the parameters of the ODE generating each observed time-series, and extrapolate the future of the ODE variables and the observations. We address this task with a variational autoencoder incorporating the known ODE function, called GOKU-net\footnote{Code available on {\urlstyle{sf} \textcolor{blue}{\url{github.com/orilinial/GOKU}}}} for Generative ODE modeling with Known Unknowns. We first validate our method on videos of single and double pendulums with unknown length or mass; we then apply it to a model of the cardiovascular system. We show that modeling the known-unknowns allows us to successfully discover clinically meaningful unobserved system parameters, leads to much better extrapolation, and enables learning using much smaller training sets.
\end{abstract}

\maketitle

\section{Introduction}
\label{sec:intro}
Many scientific fields use the language of ordinary differential equations to describe important phenomena. These include microbiology, physiology, ecology, medicine, epidemiology and finance, to name but a few. Typically, an ODE model of the form $\frac{dz(t)}{dt} = f_{\tf}(z(t))$ is derived from first principles and mechanistic understanding, where $z(t)$ are time-varying variables and $\tf$ are static parameters of the ODE model $f$. Once a model $f_{\tf}$ is specified, the values $\hat{\theta}_f$ are found that best fit every instance. These estimated values are often of great interest: in ecology these might correspond to the carrying capacity of a species, whereas in medicine they might represent the cardiovascular characteristics of a patient in the course of critical illness. Predictions based on extrapolating the estimated models are also of wide interest, for example predicting how a patient's state will evolve or respond to specific interventions. 

In many scientific applications the assumption is that the dynamic variables $z(t)$ are directly observed, possibly with some independent noise. Alternately, an assumption is often made that the observations, which we denote hereafter as $x(t)$, are a known, fixed mapping of the unobserved $z(t)$. This assumption however is not always realistic: in the case of critically-ill patients for example, while some physiological variables are directly observed such as arterial blood pressure, others that are key determinants of the dynamical system such as cardiac contractility (the heart's ability to squeeze blood), stroke volume (the volume of blood squeezed during a single heart beat), or systemic vascular resistance are not only unobserved but also have a non-trivial, and possibly unknown, mapping to the observed variables. Estimating these variables is of great clinical importance both diagnostically and in tailoring treatments aimed at their modification. 

This work addresses the scenario where, on the one hand we have the mechanistic understanding needed to define the dynamic variables $z$ and a corresponding ODE model $f_{\tf}$, but on the other hand we cannot assume that we have a good model for how the variables $z$ tie in to the observations $x$. In such a scenario, $z$ and $\tf$ take on the role of \emph{known-unknowns}: variables with a concrete meaning, which we do not know and wish to infer from data. 

Therefore, our goal is to build a learning system that can use the conjunction of mechanistic ODE models together with data-driven methods \citep{baker2018mechanistic}. Ideally this conjunction will bring out the best of both worlds, allowing us to address problems neither approach can solve on its own, especially focusing on the correct identification of these known-unknowns. Specifically, we propose an autoencoder framework called GOKU-net, standing for Generative ODE Known-Unknown net. This is a VAE architecture with the known differential equation $f$ at its heart, and with an added component that allows us to effectively use standard VAE conditional-Gaussian parameterizations, yet still obtain estimates of the known-unknown quantities which correspond to their natural physical range. 

In the next section we frame our task and its relevance to practice with an example of acute care patients. While many approaches exist for learning the parameters of ODEs, and others exist for sequence modeling with latent variables, we believe none of them can jointly address the task we outline in a straightforward way. We therefore give a (necessarily partial) overview of relevant methods and explain why we believe that despite many similarities they are not suited for the learning scenario we describe. 

In our experiments we show what gains can be made by successfully harnessing domain knowledge in terms of the ODE form: We can learn the known-unknowns, we can learn with vastly less data, and we can perform the difficult task of sequence extrapolation with much higher accuracy. 
We show this by comparing our method with several baselines such as LSTM \citep{graves2013generating} and Latent-ODE \citep{chen2018neural}\footnote{\noindent Also called Neural-ODE; we call it Latent-ODE following the usage in \citet{rubanova2019latent}.} in three domains: a video of a pendulum, a video of double pendulum, and a dynamic model of the cardiovascular system \citep{zenker2007inverse}. %We show our approach can successfully estimate known-unknowns in each case, and also outperforms methods that do not use mechanistic knowledge such as LSTM \citep{graves2013generating} and Latent-ODE \citep{chen2018neural}\footnote{\noindent Also called Neural-ODE; we call it Latent-ODE following the usage in \citep{rubanova2019latent}.} when extrapolating the observations into the future. 

    \subsection*{Our key contributions}
    %In many scenarios, an observed time series is generated by a noisy, non-linear and unknown generative process from some latent space governed by a dynamical system in the form of an ODE. We demonstrate a few in the experiments section. 
    %We demonstrate how to integrate mechanistic knowledge into a deep autoencoder framework, and show what benefits could be gained by this integration:
    \renewcommand{\theenumi}{(\arabic{enumi})}%
    \begin{enumerate}
        \item We show how to efficiently integrate mechanistic knowledge into a deep autoencoder framework.
        \item We develop experimental framework for evaluating models of ``known-unknowns'', including one based on a model of the cardiovascular-system derived from first principles. 
        \item We show what benefits could be gained by the integration of mechanistic knowledge over using purely data-driven methods:
        \begin{enumerate}
            \item Accurately inferring latent system parameters. For example, for the haemodynamically unstable patient --- estimate the relative contribution of hidden processes such as internal bleeding or reduced vascular resistance as seen in septic shock.
            \item Better extrapolation of observed time series.
            
            \item Achieving results which are equivalent to those of data-driven models while using substantially less training samples.
        \end{enumerate}
    \end{enumerate}
    
    %Specifically, we design experi
    %To this end, We introduce GOKU; a new generative approach designed to learn a probabilistic model of an observed time series that was  generated from a noisy, non-linear generative process, while utilizing a known ODE dynamical form 
    %from a that was generated from an unknown, non-trivial and noisy generative process.
    %To this end, we introduce GOKU; a new generative approach designed to utilize a known ODE system in order to learn a model of an observed time series. We demonstrate the gained benefits in three tasks: 
    %}
    %\onote{THIS FEELS REDUNDANT... DO YOU THINK WE CAN WRITE IT IN A DIFFERENT WAY?}

\section{Task definition}

We are given $N$ observed trajectories $X^i=(x^i_0, ..., x^i_{T-1})$, $i=1,\ldots N$, each describing a time evolving phenomena observed at times $t=0, \ldots T-1$. We assume each of these time sequences was generated by a noisy unknown emission process $g$ from underlying latent (unobserved) trajectories $Z^i=(z^i_0, ..., z^i_{T-1})$. The dynamics of the latent variables $Z^i$ are governed by an ODE with known functional form $f$ and unknown static parameters $\tf^i$. Note that the latent trajectories share the same functional form but have different ODE parameters across the samples $i=1,\ldots, N$:
\begin{align}
    &\frac{dz^i(t)}{dt} = f_{\theta^i_f}(z^i(t)) \label{eq.ode} 
    \\
    &x^i_t = g(z^i(t)) +  \varepsilon^i_t, \quad \varepsilon^i_t \sim \mathcal{N}(0, \sigma_{x} I) \label{eq.g}.
\end{align}
Given a training set $\{X^i\}_{i=1}^N$, and a new test sequence\\ $X'=\left(x'_0,\ldots,x'_{T-1}\right)$, our task is two-fold: 
\renewcommand{\theenumi}{(\roman{enumi})}%
\begin{enumerate}%[label=(\roman*),noitemsep,topsep=0pt]
    \item Estimate the static parameters $\tf$ for $X'$.
    %\item Estimate the latent states $z'_0,\ldots, z'_{T-1}$ corresponding to the times of the observations $X'$.
    \item Extrapolate $x'_t$ for a set of future times $t>T-1$.
\end{enumerate}

Consider the pixel-pendulum experiment we report in Section \ref{susbsec:pend}: The latent state parameter $Z$ is the pendulum's angle and angular velocity; and the ODE system $f$ is the classic pendulum equation, see Eq. \eqref{eq.pendulum}. We take the parameter $\theta_f$ to be a single number, the pendulum's length. Finally, we assume our observations $X$ are frames in a video of the pendulum, as shown in \figref{fig:pp_angle}. That means the emission function $g$ is the function that takes as input the angle of the pendulum and generates a $28 \times 28$ pixel image. The image is always scaled so that the length cannot be inferred from a single image. The task here is, given a previously unseen video, to infer the pendulum's length and to extrapolate the video into the future of the sequence. 

In the ICU patient example, $X^i$ would be a time-series of observed vital signs and other measurements such as heart rate and body temperature for patient $i$. The variables $Z^i$ describe a set of physiological variables such as blood pressure, blood volumes in the heart, cardiac stroke volumes and more. The function $f_{\tf}$ is an ODE model for these physiological variables \citep{guyton1972circulation,smith2004minimal,zenker2007inverse,ellwein2013patient,olufsen2013practical}. The parameters $\tf^i$ would be important patient-specific static variables such as arterial and venous compliances. Correct estimation of these  variables conveys immediate clinical advantage both by aiding the clinicians in establishing the correct underlying diagnosis (for example hypotension due to reduced cardiac function versus septic shock versus bleeding) and in serving as treatment goals with specific interventions tailored to the identified pathophysiological process.

We note that inferring exactly the true latent parameters might be impossible in some systems due to under-identification, as for example multiple sets of latent $z_0$ can give rise to the same observed $X$ by way of different emission models $g$.

\section{Related Work}
\begin{figure*}[ht!]
    \centering
    % \begin{subfigure}[h]{0.7\columnwidth}
    \subfigure[]{
    \includegraphics[width=0.35\textwidth]{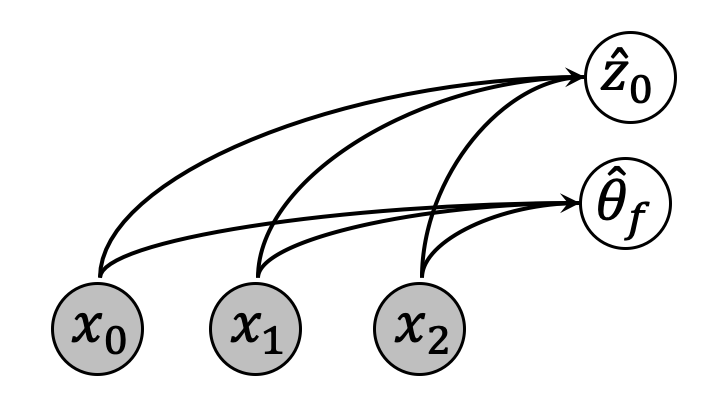}
    \label{fig:inference}
    % \caption{}
    % \end{subfigure}
    }
    \hspace{50pt}
    % \begin{subfigure}[h]{\columnwidth}
    \subfigure[]{
    \includegraphics[width=0.5\textwidth]{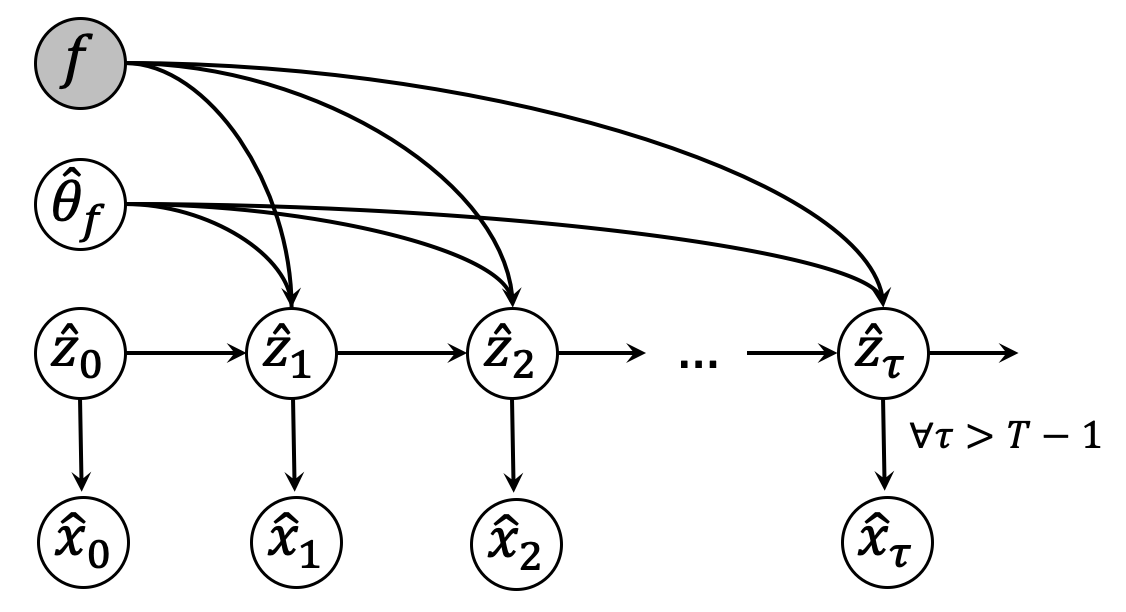}
    % \caption{}
    \label{fig:generative}
    % \end{subfigure}
    }
    \caption{\textbf{GOKU-net model.} Left: inference model; an observed signal $X_t$ is taken as input by a bi-directional LSTM to produce estimates of the initial state $\hat{z}_0$ and static parameter $\hat{\theta}_f$. Right: The ODE solver takes these values together with the given ODE function $f$ to produce the latent signal $\hat{Z}$, then reconstruct $\hat{X}$ using an emission network $\hat{g}$. The ODE solver can integrate $\hat{z}_t$ arbitrarily far forward in time, enabling the extrapolation of $X$ for any $t=T+\tau$.}
    \label{fig:model}
\end{figure*}

There has been much work recently bringing together ML methods and ODEs. However, we believe we are unique in our setting, as explained below and outlined in Table \ref{tab:related_work}.
We divide existing work into several categories: First, work on parameter identification in dynamical systems which assumes both the ODE function $f$ and the emission function $g$ are known. Second, work on latent state sequence modeling. This work does not assume any known dynamics or emission model. Finally, there is recent work tying together machine learning models and physical models in a task-specific way.
%We summed up the differences between some of these methods and ours in \cref{tab:related_work}.

%\textbf{Methods for parameter identification }
ODE parameter identification has been the subject of decades of research across many scientific communities. 
Classic state-space models, including methods such as the Kalman filter \citep{kalman1960contributions} and its non-linear extensions \citep{jazwinski2007stochastic,julier1997new,wan2000unscented}, can learn the parameters of a dynamic system from observations; however, they are limited to the case where the emission function $g$ is known. Moreover, they usually perform inference on each sequence separately. 
Many machine learning methods have been proposed for this task, for example using reproducing kernel Hilbert space methods \citep{gonzalez2014reproducing} and Gaussian Processes \citep{dondelinger2013ode,barber2014gaussian,gorbach2017scalable}, Fast Gaussian Process Based Gradient Matching \citep{wenk2018fast} and recent follow up work \citep{wenk2019odin}.
In general these methods assume in practice that the observed signal is simply the state variable $z(t)$ with independent additive noise, whereas we allow a more complicated, and apriori unknown, link from latent states to observations. %In our notation that would imply $x_t = z_t + \varepsilon_t$, where $\varepsilon_t \sim \mathcal{N}(0,\sigma^2 I)$.
%If, in addition to the ODE function $f$, we assume the generative function $g$ is also known, standard filtering approaches such as the unscented Kalman filter \citep{julier1997new,wan2000unscented} could succeed in all tasks above. A large number of recent papers dealt specifically with the case where $g$ is additive noise XXX citeXXX. In Section \ref{sec:related} below we outline XXX

%\textbf{Sequence modeling }
Many methods for extrapolation of a given signal assuming there is some unknown arbitrary underlying latent sequence have been proposed. Some prominent examples are LSTM \cite{graves2013generating}, Deep Markov Models \citep{krishnan2017structured}, Neural ODE (also called Latent-ODE, L-ODE, \citep{chen2018neural}) and follow up methods, \citep{rubanova2019latent,yildiz2019ode2vae},
NbedDyn \citep{ouala2019learning}, the work by \citet{ayed2019learning}, the Disentangled State Space Model (DSSM, \citep{miladinovic2019disentangled}), SVG-LP \citep{denton2018stochastic}, method using Gaussian Processes \citep{heinonen2018learning}, and Gaussian Process State Space Models (GPSSM, \citep{eleftheriadis2017identification,zhao2020state}); motivated by robotics applications, \citet{watter2015embed} aim to learn latent space such that locally linear dynamics will be useful. These methods do not infer the ODE parameters as they do not learn any intrinsically meaningful latent space. They also do not exploit the prior information embedded in the mechanistic knowledge underlying the derivation of the ODE system $f$. Of the above methods, DSSM has been shown to learn a latent space which might under the right circumstances correspond to meaningful parameters, but that is not guaranteed, nor is it the goal of the method. In the healthcare regime,  \citet{cheng2019patient} proposed a method for learning a sequence which includes a dynamic system in the form of a latent force model \citep{alvarez2009latent}; this approach builds on learning to fit general basis functions to describe the observed dynamics, and does not take as input an ODE system derived from prior mechanistic understanding.

%\textbf{Machine learning with mechanistic components }
Closer in spirit to our work is the work by Greydanus et al. \citep{greydanus2019hamiltonian} on Hamiltonian neural networks. In their model the latent space can be interpreted in the form of learning a conserved physical quantity (Hamiltonian). Although related to our work, we note that not all ODE systems have easily identified conserved quantities. Specifically, the systems that motivate our research do not usually have a Hamiltonian representation. For example in section \ref{sec:exp} we experiment with an ODE system of a pendulum with friction as a non-conservative quantity. Another closely related work is SINDy Autoencoders \citep{champion2019data}. This work is similar to ours in using an unknown and complex emission function, and learning a physically meaningful latent space. Their model is not given the ODE system $f$, but instead aims to learn some representation of the underlying ODE as a linear combination of bases functions of the latent state. %This work can be used in many tasks where the latent space follows an ODE, but there is no reason to believe it can discover, on its own, highly involved dynamics such as those of the cardiovascular system. E.g., in a model of the cardiovascular system we use in \cref{sec:exp}, the ODE system is very complex, and the ODE parameters are integrated in the ODE in a non trivial way.
In the field of learning for healthcare,  \citet{soleimani2017treatment} show how a specific ODE model, the linear time-invariant impulse-response model, can be used in conjunction with latent-space models to estimate how a patient's measurements would react to interventions. It brings together mechanistic modeling in terms of response to impulse treatments, along with data-driven modeling using Gaussian processes. It does not learn emission functions from the latent space, and focuses on the specific ODE model relevant to their task.

\begin{table*}[h]
    \centering
    \begin{tabular}{lc C{25mm} C{30mm} C{25mm}} 
        \toprule
        Method & ODE function  & Emission function & $\tf$ identification & $X$ extrapolation \\
        \midrule
        
        LSTM  \scriptsize{\citep{graves2013generating}} & not required & learned & \xmark  & \cmark\\  
        
        L-ODE \scriptsize{\citep{chen2018neural,rubanova2019latent}} & not required & learned & \xmark  & \cmark\\
        
        DMM
        \scriptsize{\citep{krishnan2017structured}} & not required & learned & \xmark  & \cmark \\
        
        GPSSM
        \scriptsize{\citep{eleftheriadis2017identification}} & not required & learned & \xmark  & \cmark \\

        DSSM \scriptsize{\citep{miladinovic2019disentangled}} & not required & learned & \xmark  & \cmark\\

        HNN \scriptsize{\citep{greydanus2019hamiltonian}} & can be used & learned & \xmark  & \cmark\\ 
        
        NbedDyn \scriptsize{\citep{ouala2019learning}} & not required & partially given & \xmark  & \cmark \\
        
        ODIN \scriptsize{\citep{wenk2019odin}} & required & given & \cmark  & \xmark \\
        
        UKF  \scriptsize{\citep{wan2000unscented}} &  required & given & \cmark  & \cmark\\  
        
        GOKU-net & required & learned & \cmark  & \cmark \\
    \bottomrule
    \end{tabular}
    \caption{Related Work: for each method we indicate does it require the ODE function $f$ as input; does it assume the emission function $g$ is given or learned; does it allow identification of the ``known-unknown'' parameters $\theta_f$; and whether it allows for extrapolating the observed $X$.}
    \label{tab:related_work}
\end{table*}

\section{Model and method}\label{sec:meth}
Given $N$ observed trajectories $X^i=(x^i_0, ..., x^i_{T-1})$, $i=1,\ldots N$, our main idea is based on inferring a latent trajectory $\hat{Z}$ while reconstructing the ODE parameters $\tf$, in a variational autoencoder approach \citep{rezende2014stochastic,kingma2013auto}. As usual, that implies learning both an inference function (encoder) and an emission function (decoder). 
The inference function takes an observed sequence $X^i$ as input, and has two components: The first infers the ODE parameters $\htf$, and the second infers an initial latent state $\htz$. We next use the \emph{known} ODE functional form $f$, the inferred ODE parameters $\htf$ and the inferred initial state $\htz$ to obtain an estimated trajectory $\hat{Z}^i$ by a numerical ODE solver. We then use $\hat{Z}^i$ as input to a \emph{learned} emission function $\hat{g}$, obtaining a reconstructed signal $\hat{X}^i$. We estimate the log-likelihood of the reconstructed signal, and use stochastic backpropagation \citep{rezende2014stochastic,kingma2013auto} through the ODE solver in order to update the parameters of the inference network and emission model -- details below. Extrapolating the latent trajectory using the ODE solver lets us make estimates of $X^i$ arbitrarily far forwards or backwards in time. \figref{fig:model} illustrates the proposed model.

\subsection{Generative model and inference}
Using the relationships between latent and observed variables given in Eqs. \eqref{eq.ode} and \eqref{eq.g}, we define a generative model over the set of ODE parameters $\tf$, the latent states $Z$, and the observations $X$. While we assume the true ODE function $f$ is given, we estimate the emission by a learned function $\hat{g}$. 

A crucial issue we must address is that in standard VAEs the prior distributions of the latent vectors $z_0$ and $\tf$ are set to be a zero-mean unit-variance Gaussian. However, we are conducting inference in a latent space where the variables correspond to  specific variables with physical meaning and constraints: for example, the variable for blood volume has a limited set of realistic values.
We overcome this by defining arbitrary (non-physical) latent vectors $\tz$ and $\ttf$ with standard Gaussian priors, and then learning deterministic ``physically-grounding'' transformations $h_z$ and $h_{\tf}$ such that: 
\begin{align}
    \tz \sim \mathcal{N}(0, I), \quad z_0 &= h_z(\tz)  \label{eq.hz},\\
    \ttf \sim \mathcal{N}(0, I), \quad \theta_f &= h_{\theta} (\ttf)  \label{eq.htf}.
\end{align}
We then have $z_t$, $t>0$ and $X$ generated following Eqs. \eqref{eq.ode} and \eqref{eq.g}.
With the above generative model, we have the following factorized joint distribution over latent and observed variables:
\begin{align*}
   p(X, Z, \tf, \tz, \ttf) = 
    & p(\tz)  p(\ttf)  p(z_0 | \tz)  p(\tf | \ttf)
    p(x_0 | z_0)\\
    &\prod_{t=1}^{T-1} p(x_t|z_t)p(z_t|z_{t-1}, \tf).
\end{align*}
This follows due to the conditional independence: for $t' \neq t$: $x_t \independent (z_{t'}, \tf, \ttf, \tz) | z_t$ and for $t' \neq t-1$: $z_t \independent (X, \ttf, \tz, z_{t'}) | z_{t-1}, \tf$. The probabilities $p(z_0 | \tz)$ and $p(\tf | \ttf)$ are deterministic, meaning they are Dirac functions with the peak defined by Eqs. \eqref{eq.hz} and \eqref{eq.htf}.
The transition distribution $p(z_t|z_{t-1}, \tf)$ is also a Dirac function with the peak defined by Eq. \eqref{eq.ode}. Finally, the emission distribution $p(x_t|z_t)$ is defined by Eq. \eqref{eq.g}. 

\textbf{Inference} We define the following joint posterior distribution over the unobserved random variables $Z$, $\tf$, $\tz$ and $\ttf$, conditioned on a sequence of observations $X$:
\begin{align*}
    q(Z, \tf, \tz, \ttf | X) 
    = 
    &q(\tz | X)   q(\ttf | X)   q(z_0 | \tz)   q(\tf | \ttf)\\
    &\prod_{t=1}^{T-1} q(z_t | z_{t-1}, \tf).
\end{align*}
The inference network conditionals $q(z_t| z_{t-1}, \tf)$, \hspace{1mm} $q(z_0 | \tz)$ and $q(\tf | \ttf)$ are deterministic and mirror the generative model as defined in Eqs. \eqref{eq.ode}, \eqref{eq.hz} and \eqref{eq.htf},
respectively. 
For the posteriors $q(\tz| X)$ and $q(\ttf | X)$ we use conditional normal distributions where $\phi^{enc}_{\tz}$ and $\phi^{enc}_{\ttf}$ are learned NNs:
\begin{align*}
    &q(\tz|X) = \mathcal{N}(\mu_{\tz} , \sigma_{\tz}),
    \quad [\mu_{\tz}, \sigma_{\tz}] = \phi^{enc}_{\tz}(X),
    \\
    &q(\ttf|X) = \mathcal{N}(\mu_{\ttf} , \sigma_{\ttf}),
    \quad [\mu_{\ttf}, \sigma_{\ttf}] = \phi^{enc}_{\ttf}(X).
\end{align*}

\subsection{Objective}
We use the evidence lower-bound (ELBO) variational objective \citep{kingma2013auto,rezende2014stochastic}:
\begin{align} 
    \mathcal{L}(X)
    = 
    & \mathbb{E}_{q(Z, \tf, \ttf, \tz | X)} \! \left[\log p(X | Z, \tf, \tz, \ttf) \right] \nonumber
    - \\ 
    & KL \! \left[ q(Z, \tf, \tz, \ttf | X) ||  p(Z, \tf, \tz, \ttf)     \right] \! .
    \label{eq:obj}
\end{align}

Since for all $t' \neq t$ we have $x_t \independent (x_{t'}, z_{t'}, \tf, \tz, \ttf)| z_t$, the first term of Eq. \eqref{eq:obj} decomposes as: 
\begin{align*}
 &\mathbb{E}_{q(Z, \tf, \ttf, \tz | X)} \left[\log p(X | Z, \tf, \tz, \ttf) \right] =\\ 
    &\sum_{t=0}^{T-1} \mathbb{E}_{q(z_t | X)} \left[\log p(x_t | z_t) \right].
\end{align*}
The KL term decomposes into the following sum of KL terms:
\begin{align*}
    &KL \left[ q(Z, \tf, \tz, \ttf | X) ||  p(Z, \tf, \tz, \ttf)     \right] = \\
     &KL \Big[   q(\ttf| X) ||  p(\ttf)       \Big]
    + KL \Big[   q(\tz| X) ||  p(\tz)       \Big].
\end{align*}
See the appendix for the full derivation.

\subsection{Implementation}
We model $[\hat{g}, h_z, h_{\tf}, \phi^{enc}_{\tz}, \phi^{enc}_{\ttf}]$ as neural networks: 
 $\hat{g}$, $h_z$ and  $h_{\tf}$ as fully connected neural networks; $\phi^{enc}_{\tz}$ as an RNN which goes over the observed $X$ backwards in time to predict $z_0$; and $\phi^{enc}_{\ttf}$ as a bi-directional LSTM \citep{huang2015bidirectional} with fully connected networks from $X$ into $\ttf$. We use bi-LSTM for $\tf$ identification since $\tf$ is time invariant.

In order to perform stochastic backpropagation, we must calculate the gradient through the ODE defined by $f$ and $\theta_f$.
To this end, we experimented with two methods: an explicit approach, where the gradients are propagated directly through the operations of the numerical ODE integration, and an implicit approach\footnote{https://github.com/rtqichen/torchdiffeq} that uses the adjoint method which is more memory efficient \citep{chen2018neural}. In both methods we used the Runge-Kutta-4 numerical integration. Both methods achieved very similar results, therefore we present in this paper only the latter.

\section{Experiments}\label{sec:exp}
In this section we analyze how GOKU-net can be used for observed signal extrapolation and ODE parameter identification. We first validate the method using two widely known physics models: an OpenAI Gym video simulator of a pendulum \citep{brockman2016openai}, introduced as an ODE modeling benchmark in \citep{greydanus2019hamiltonian}, and a similar OpenAI Gym video simulator of a (chaotic) double pendulum. We then apply the method to a model of the cardiovascular system based on the model introduced in \citep{zenker2007inverse}.
In each case we train the model on a set of sequences with varying ODE parameters ($\tf$) and initial conditions ($z_0$), and test on unseen sequences with parameters and initial conditions sampled from the same distribution as the train set. 
Each dataset was randomly divided into train, validation and test sets (80\%, 10\%, 10\%). Validation was used for early stopping, and hyper-parameter tuning. %We evaluate the methods across three grounding conditions: no grounding latent states $Z$ (denoted $0\%$), and access to randomly sampled $1\%$ or $5\%$ of the latent $Z$. Grounding observations are available only during training, except for the DI baseline which must use them at test time too, see below. We sometimes denote grounded methods by the percent of grounding, e.g. GOKU-1\%. In the appendix we give the full details of the architectures and setup of each method and dataset.

%Many of the systems we are interested in suffer from  non-identifiability of the ODE parameters under the true generative model -- different sets of parameters can give rise to the exact same observations. We therefore also experiment with giving the models access to a sparse set ( $1\%$ or $5\%$) of observations of the true latent variables $Z$, denoted $Z^{\text{observed}}$. we now describe. 

\subsection{Baselines}\label{subsec:base}
 
\tabref{tab:baslines} summarizes what type of information each of the baselines described below has access to.

\textbf{Data-driven }
As baselines with no input from mechanistic models we use (i) LSTM  \citep{graves2013generating}, (ii) Latent-ODE \citep{chen2018neural}, originally called Neural-ODE; we denote it L-ODE, (iii) DMM \citep{krishnan2017structured}. These methods can only be used to extrapolate $X$ for future time steps, since their latent space has no physical interpretation.

\textbf{Direct inference (DI) }
We introduce a baseline which can use both the known model $f$ and the observations $X$ by having access to an additional bit of knowledge: sparse observations from the latent states $Z$, during both train and test time. Note that none of the other baselines, nor GOKU, has access to such information. 

We call this baseline ``direct inference'' (DI). It has two steps: First inferring the ODE parameters and latent states, and then, separately, learning the connection between the latent states $Z$ and observations $X$.

Specifically, the ODE parameters and latent states $(\hat{Z}, \hat{\theta}_f)$ are estimated directly from a sparse sequence of directly observed latent states $Z^{\text{observed}}$, by minimizing the loss function:
\begin{equation}
    \mathcal{L}_{ground} = \sum_{t} M(t) \cdot \Vert  \hat{z}_t - z^{\text{observed}}_t  \Vert_2^2,
    \label{eq.ground_loss}
\end{equation}
where $M(t)\in \{0,1\}$ indicates for which time points the latent variables are observed. $\hat{z}_t$ is the latent vector predicted by the model and $z^{observed}_t$ are the observed samples of the latent vectors.  We optimize Eq. \eqref{eq.ground_loss} using gradient descent through an ODE solver with the given ODE model $f$; details in the appendix. 
Then in the second step, DI obtains estimates for the observations $X$ by constructing a training set where the instances are the $Z$ inferred for each training sequence, and the labels are the corresponding observations $X$. We then learn a function $\hat{g}$ predicting $X$ from $Z$. Below we denote the method as DI-q\%, where $q\in\{1,5\}$ indicates what percent of the latent $Z$ were available to DI during train and test. We emphasize that this baseline is the only one to have access to observations of $Z$, and that these are available to it both at train and test time.

\begin{table}
\centering
    \begin{tabular}{lccc} 
        \toprule
        Method & $f$  & $Z$ \\
        \midrule
        
        LSTM  \citep{graves2013generating} & \xmark &  \xmark \\  
        
        Latent-ODE \citep{chen2018neural} & \xmark  & \xmark \\  
        
        DMM \citep{krishnan2017structured} & \xmark  & \xmark \\  

        %Latent-ODE+ & \xmark & \cmark \\  
        
        DI  & \cmark &  \cmark $^*$ \\  
         
        GOKU-net & \cmark &  \xmark \\  
        
        %GOKU-q\%, \footnotesize{$q>0$} & \cmark &  \cmark \\  
        
         HNN \citep{greydanus2019hamiltonian} & \cmark  & \xmark \\
    \bottomrule
    \end{tabular}
    \caption{Each of the methods we compare, and what information does it have access to beyond the observations $X$: the true ODE function $f$, and sparse grounding latent states $Z$. $^*$DI has access to $Z$ at test time as well as train.}
    \label{tab:baslines}
\end{table}

\subsection{Single Pendulum From Pixels}\label{susbsec:pend}
In this experiment we generate a video of a pendulum, aiming to predict the future frames and to identify its parameter.
We use the following non-linear oscillator ODE:
\begin{align}
    \frac{d\theta(t)}{dt} = \omega(t), \quad \quad
    \frac{d\omega(t)}{dt} = -\frac{g}{l} \sin{\theta(t)},
    \label{eq.pendulum}
\end{align}
with gravitational constant $g=10$. The ODE has a single parameter which is the pendulum's length $l$, and the ODE state is $z_{t} = \left(\theta(t), \omega(t)\right)$. %This task is more challenging than the Lotka-Volterra one above, because of the complex emission function we used, as we explain next.

\textbf{Data Set}
We followed \citep{greydanus2019hamiltonian} and used the \verb|Pendulum-v0| environment from OpenAI Gym \citep{brockman2016openai}. For training we simulated 500 sequences of 50 time points, with time steps of $\Delta t = 0.05$ and pre-processed the observed data such that each frame is of size $28 \times 28$. We made one important change relative to \citep{greydanus2019hamiltonian}: the ODE parameter $l$ was uniformly sampled, $l\sim U[1, 2]$ instead of being constant, making the task much harder. The images are scaled so that the length of the pendulum is not identifiable from a single frame.
Each test set sequence is 100 time steps long, where the first 50 time steps are given as input, and the following 50 were used only for evaluating the signals extrapolation.

%As in the Lotka-Volterra experiment, we compare between our method and the baselines on the pixel-pendulum data set. 

%\unote{We note that a comparison to HNN \cite{greydanus2019hamiltonian} is not relevant since they used a single parameter which is a much easier task. }

%%%%%%%%%%%%%%%%%%%%%%%%%%%%%%%
%%%%% Update for AAAI, original in NeurIPS submission
%%%%%%%%%%%%%%%%%%%%%%%%%%%%%%%
% \begin{table}
%     \centering
%     \begin{tabular}{lccc} 
%         \toprule
%         Method & DI 5\% & DI 1\% & GOKU-net \\
%         \midrule
%          $\hat{\theta}_f$ error & 0.077 $\pm$ .029 & 0.511 $\pm$ .044 &  0.078 $\pm$ .008  \\ 
%         \bottomrule
%     \end{tabular}
%     \label{tab:pp.params}
%     \caption{Pixel-pendulum mean $L_1$ error for predicting $\tf$ across test samples, with standard error of the mean. DI $5\%$ and DI $1\%$ indicate how many of the $Z$ latent states were observed by DI (direct inference). Methods not presented cannot perform the given task.} 
%     \label{tab:pp.results}
% \end{table}

\begin{table}[h]
    \centering
    \begin{tabular}{ccccc} 
        \toprule
        Method & X extrap. & $\|\tf - \hat{\theta}_f\|_1$ & $corr\left(\tf,\hat{\theta}_f\right)$  \\
        \midrule
        % GOKU-net & 7 $\pm$ 1& 78$\pm$8\\
        % Latent-ODE & 42 $\pm$ 2 & N/A\\
        % LSTM & 81 $\pm$ 4 & N/A\\
        % DI 1\% & 147 $\pm$ 0 & 77$\pm$29\\
        % DI 5\% & 147 $\pm$ 0 & 511$\pm$44\\
        GOKU-net & 7 $\pm$ 1& 78 $\pm$ 8 &0.979\\
        Latent-ODE & 42 $\pm$ 2 & N/A  & N/A\\
        LSTM & 81 $\pm$ 4 & N/A & N/A\\
        DMM & 74 $\pm$ 5 & N/A & N/A\\
        DI 1\% & 147 $\pm$ 0 & 511$\pm$44 & 0.046\\
        DI 5\% & 147 $\pm$ 0 & 77$\pm$29 & 0.751 \\
    \bottomrule
    \end{tabular}
    \caption{Pixel pendulum mean $L_1$ error ($\times 10^3$) for extrapolating $X$ with standard error of the mean, and $\tf$ identification error (both $L_1$ and correlation coefficient) across test samples. DI $5\%$ and DI $1\%$ indicate the percentage of the $Z$ latent states were observed by DI. The X extrapolation $L_1$ error of simply predicting all pixels are zero is 147, meaning both DI baselines failed in this task.} 
    \label{tab:pp.results}
\end{table}

\begin{figure}[h]
    \centering
    \includegraphics[width=\columnwidth]{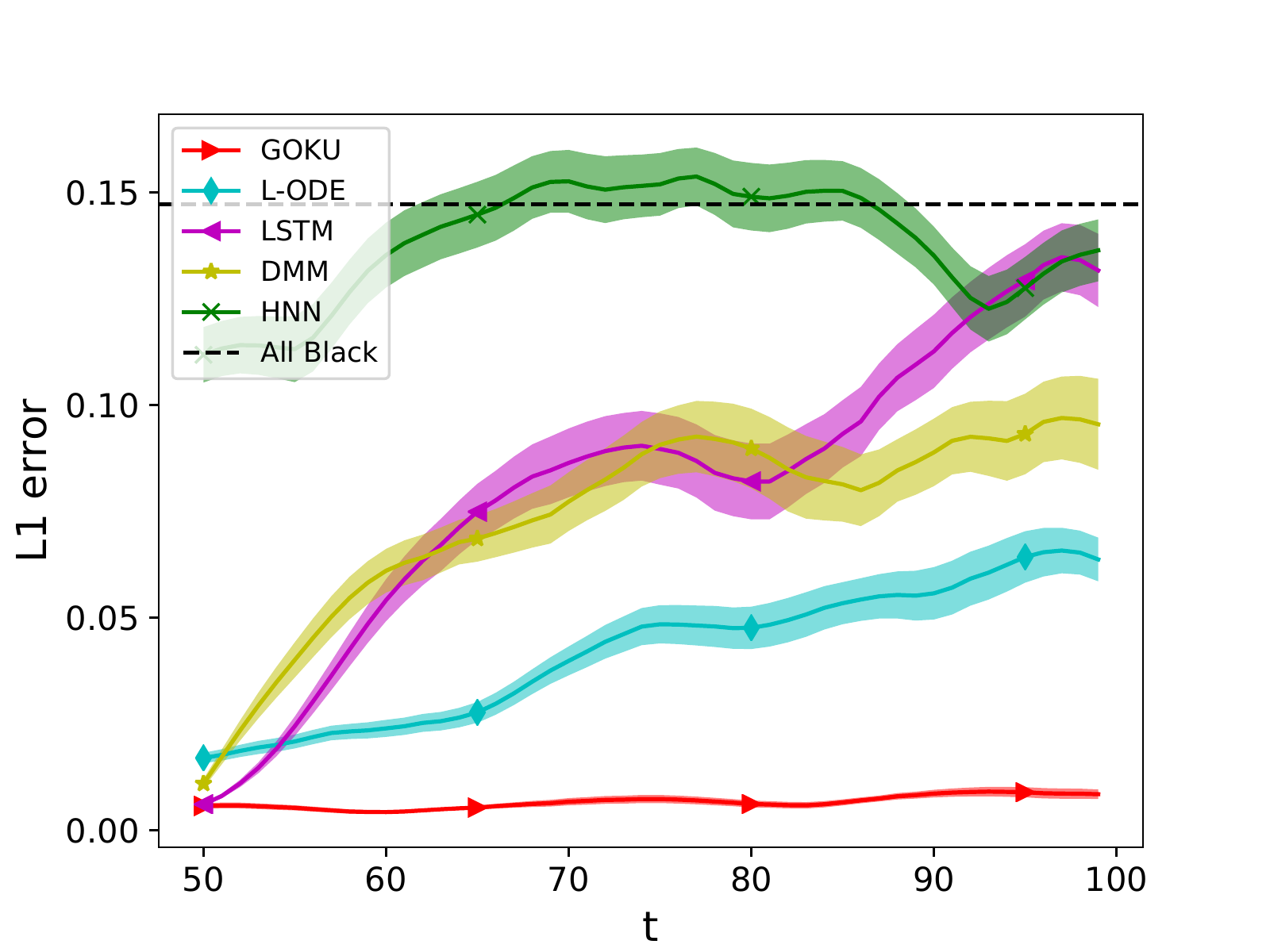}
    \caption{Pixel pendulum: mean extrapolation error for observations $X$ over time steps after end of input sequence. Percentages in legend are percent grounding observation in training. HNN by \citet{greydanus2019hamiltonian}. ``All black'' is predicting all pixels as zero brightness (black).}
    \label{fig:pp_x_pred}
\end{figure}

\begin{figure}[h]
    \centering
    \includegraphics[width=\columnwidth]{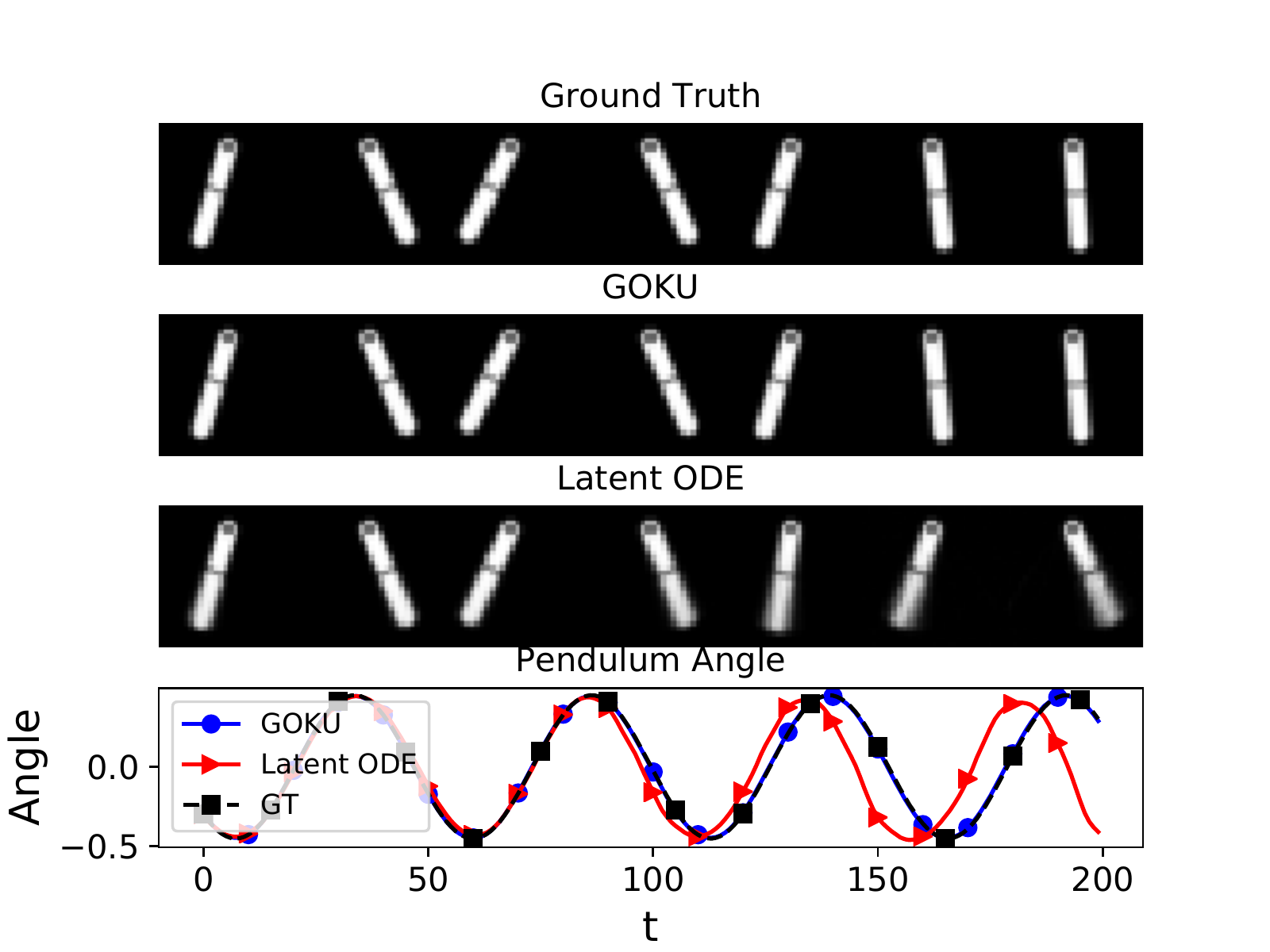}
    \caption{Predicting the dynamics of pixel pendulum. The first 50 frames are observed, and the next 150 are predicted. The above 3 figures are examples of each method's predicted frames every 30 time steps.}
    \label{fig:pp_angle}
\end{figure}

%%%%%%%%%%%%%%%%%%%%%%%%%%%%%%%
\textbf{Evaluation and results.} \tabref{tab:pp.results} shows the ODE length parameter identification error in terms of $L_1$ error and the Pearson correlation coefficient $r$, and mean extrapolation error in terms of $L_1$ error ($L_2$ has similar results). \figref{fig:pp_x_pred} shows how the extrapolation error of the observed signals evolves over time, starting from time $t=50$. In \figref{fig:pp_angle} we demonstrate how GOKU-net extrapolates on a single, randomly selected, signal when compared to the best performing baseline, the Latent ODE.

In terms of identification, GOKU-net performs much better than the DI 1\% baseline, and similar to the DI 5 \% baseline, even though GOKU-net does not receive access to observations of $Z$, which are accessible to the DI baselines. When performing extrapolation, we see in \figref{fig:pp_x_pred} that GOKU-net extrapolates much better than all baselines. Specifically, GOKU-net outperforms HNN \citep{greydanus2019hamiltonian} which has difficulty with the fact that the ODE parameter is not constant.

\begin{figure}[h]
    \centering
    \includegraphics[width=\columnwidth]{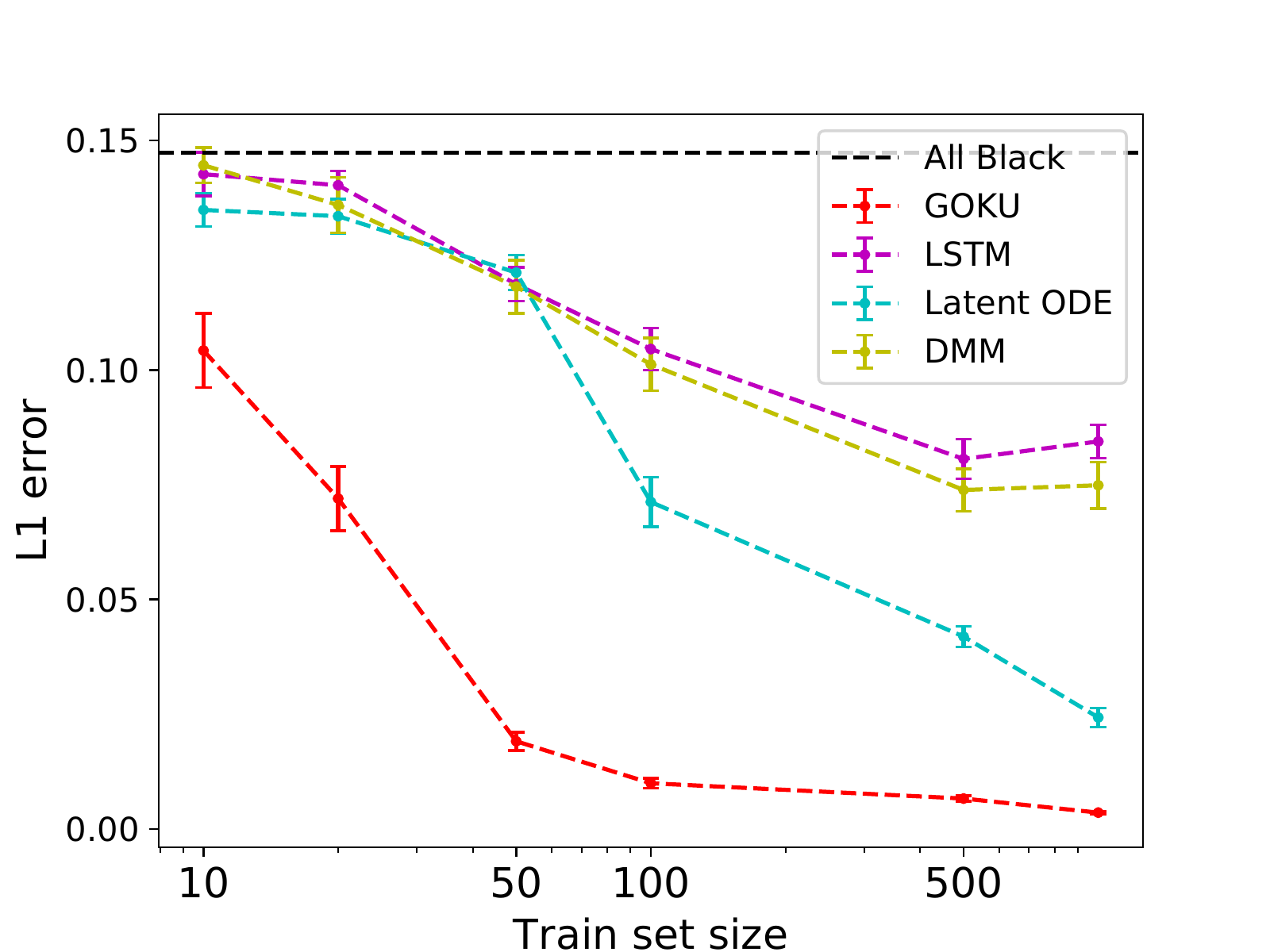}
    \caption{Pixel pendulum X extrapolation error on test set, for different train set sizes. ``All black'' is simply predicting all pixels as zero brightness (black).}
    \label{fig:pp_train_size}
\end{figure}

We further tested performance as a function of train set size, reasoning that domain knowledge should help reduce the need for many training samples. We evaluated the $L_1$ extrapolation error of the different methods averaged over time steps $t=50 \ldots 100$, taken on the same test set. As can be seen in \figref{fig:pp_train_size}, training GOKU-net on as little as 50 train samples provides better X-extrapolation error than Latent ODE achieves with 1000 (20 fold) train samples, showing we can achieve substantial benefits by properly using domain knowledge. Both DI baselines completely failed in this scenario.

%\subsubsection{Train set size reduction}
%One of the major problems in machine learning, is the need of a lot of train samples. This is a problem both because the data may be hard to gather, and because training models on big dataset is computationally expensive. We believe that using the added knowledge in for form of known ODE could reduce the needed training data substantially. To show this, we trained GOKU-net, LSTM and Latent ODE on different sizes of train sets, and evaluated them on the same test set. As can be seen in \cref{fig:pp_train_size}, training GOKU-net on as little as 50 train samples provides better X-extrapolation error than Latent ODE achieves with 1000 (20 fold) train samples.

\subsubsection{Pixel Pendulum with Unknown Unknowns}\label{subsubsec:uu}
\begin{figure}[h]
    \centering
    \includegraphics[width=\columnwidth]{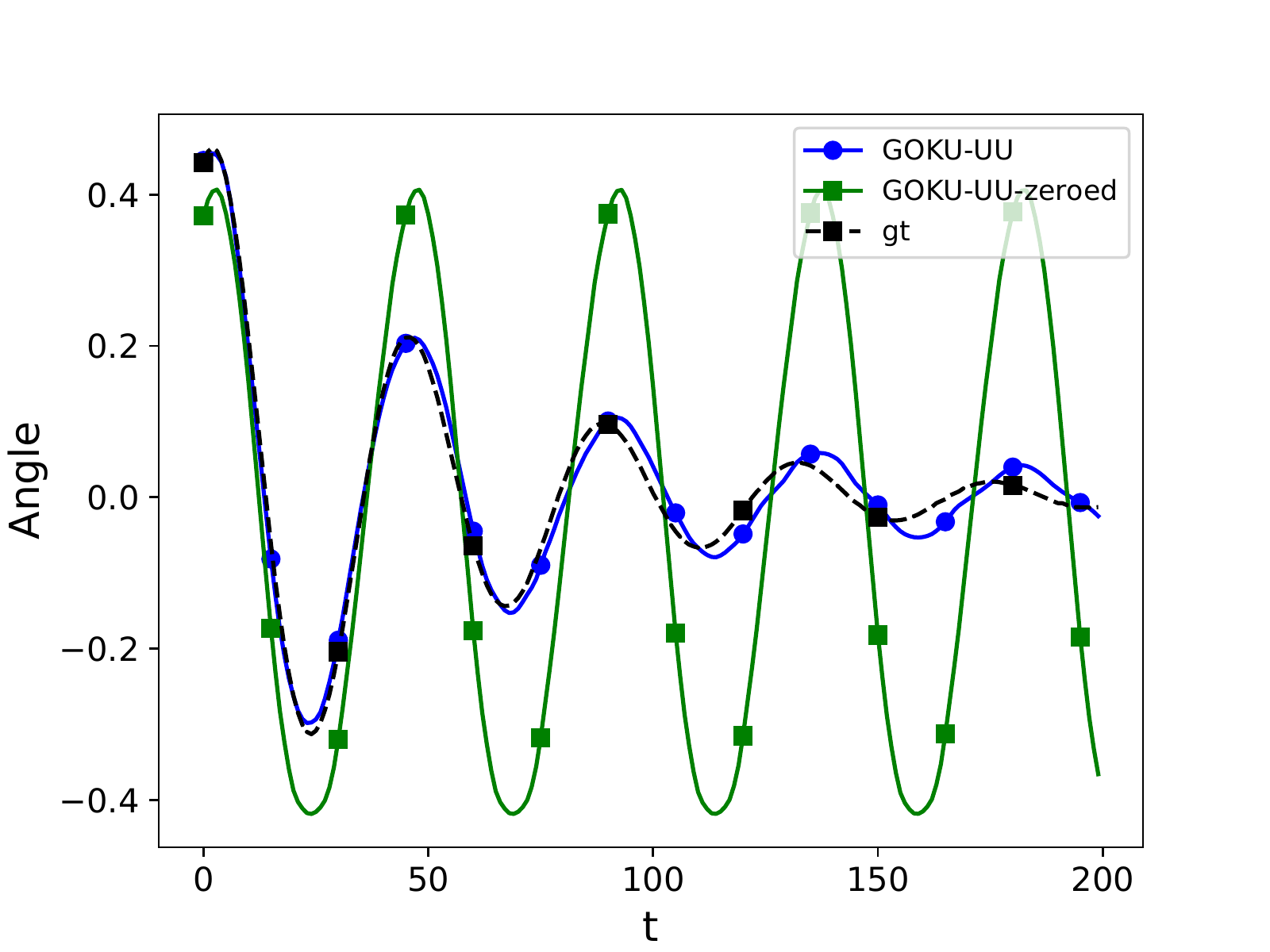}
    \caption{Pixel pendulum with friction predicted angle example. Here we demonstrate that zeroing the $f_{abs}$ part of GOKU-UU, results in a friction-less signal.}
    \label{fig:pp_friction_angle_zeroed}
\end{figure}

In this experiment we relax an important assumption we made so far: that we have the full ODE model $f$ for the underlying latent states. Instead, here we experiment with data generated by a true ODE model $f$ that is different from the one given as input to GOKU-net. We do this by adding friction to the pendulum ODE Eq. \eqref{eq.pendulum}:
\begin{align*}
   \frac{d\theta(t)}{dt} = \omega(t), \quad \quad
    \frac{d\omega(t)}{dt} = -\frac{g}{l} \sin{\theta(t)} -\frac{b}{m}\omega(t).
\end{align*}
Importantly, GOKU-net is still given the ODE defined by Eq. \eqref{eq.pendulum}, without friction. For this task we added a trainable abstract function $f_{abs}$ to the ODE function such that now $\frac{dz_t}{dt} = f_{ODE}(z_t, \tf) + f_{abs}(z_t, \tf)$, where $f_{ODE}$ is the given friction-less ODE Eq. \eqref{eq.pendulum}. The idea is that the abstract function $f_{abs}$ might model the \emph{unknown-unknowns} of the system, in this case the friction. We name this approach GOKU-UU for GOKU with unknown-unknowns. The results show that even with the incorrect ODE model, we were able to extrapolate the signal including the decay of the pendulum's velocity. We further demonstrate in \figref{fig:pp_friction_angle_zeroed}, that when we zero the inferred $f_{abs}$ at test time the inferred pendulum's velocity did \emph{not} decay, strengthening the claim that $f_{abs}$ modeled the friction, while $f_{ODE}$ modeled the friction-less pendulum. Results are presented in the appendix, as well as algorithmic details.

\subsection{Double Pendulum From Pixels}
In order to further test the limits of GOKU, we created a dataset comprised of videos of a double pendulum. The dynamics of the double pendulum are much more complex than those of the single pendulum, and are known to have chaotic nature \cite{shinbrot1992chaos}, making this task a much harder one. The dataset creation was similar to the single pendulum experiment, using the mass of the second pendulum as the unknown parameter $\theta_f$ we wish to identify. More details about the ODE system, and the dataset creation process in the appendix.\\

\tabref{tab:DP.results} shows the ODE parameters identification error in terms of the (Pearson) correlation coefficient and $L_1$ error, and mean extrapolation error in terms of $L_1$ error. Although the double pendulum has a chaotic nature, GOKU-net was still able to extrapolate the observed video signals well into the future, and better than the compared baselines. In terms of identification of the mass parameter, GOKU-net managed to correlate very well with the ground truth, but was off in its scale, predicting masses which were too small.

\begin{table}[h]
    \centering
    \begin{tabular}{ccccc} 
        \toprule
        Method & X extrap. & $\|\tf - \hat{\theta}_f\|_1$ & $corr\left(\tf,\hat{\theta}_f\right)$  \\
        \midrule
        GOKU-net & 12 $\pm$ 0& 1.378 $\pm$ 0.030 & 0.972\\
        Latent-ODE & 16 $\pm$ 1 & N/A& N/A\\
        LSTM & 32 $\pm$ 1 & N/A& N/A\\
        DMM & 26 $\pm$ 1 & N/A& N/A\\
        DI 1\% & 37 $\pm$ 0 & 0.261 $\pm$ 0.041 & 0.536\\
        DI 5\% & 37 $\pm$ 0 & 0.057 $\pm$ 0.005 & 0.998\\
    \bottomrule
    \end{tabular}
    \caption{Double pendulum mean $L_1$ error ($\times 10^3$) for extrapolating $X$ with standard error of the mean, and $\tf$ identification error (both $L_1$ and correlation coefficient) across test samples. DI $5\%$ and DI $1\%$ indicate the percentage of the $Z$ latent states were observed by DI. The X extrapolation $L_1$ error of simply predicting all pixels are zero is 37, meaning both DI baselines failed in this task.}
    \label{tab:DP.results}
\end{table}

\subsection{Cardiovascular System}
As mentioned in the introduction, a major clinical challenge in the context of critical care lies in elucidating the causes for haemodynamic instability and deriving an appropriate treatment plan. We approached this challenge using a simplified mechanistic model of the cardiovascular system (CVS), which is a modified version of a model first suggested in \citet{zenker2007inverse}. It is a multi-compartment model comprising the heart, the venous and arterial subsystems together with a reflex loop component representing the nervous system control of blood pressure:
\begin{align*}
    % dSV/dt
    \frac{dSV(t)}{dt} &= I_{external},  \\
    % dPa/dt
    \frac{dP_a(t)}{dt} &= \frac{1}{C_a}\left(\frac{P_a(t) - P_v(t)}{R_{TPR}(S)} - SV\cdot f_{HR}(S)\right) \\
    % dPv/dt
    \frac{dP_v(t)}{dt} &= \frac{1}{C_v}\left(-C_a\frac{dP_a(t)}{dt} + I_{external} \right)\\
    % dS/dt
    \frac{dS(t)}{dt} &= \frac{1}{\tau_{Baro}}
    \left(1 - \frac{1}{1 + e^{-k_{width}(P_a(t) - P_{a_{set}})}   } - S  \right),
\end{align*}
where 
\begin{align*}
    R_{TPR}(S) &= S(t)(R_{TPR_{Max}} - R_{TPR_{Min}}) 
    + R_{TPR_{Min}} + R_{TPR_{Mod}}\\
    f_{HR}(S) &= S(t)(f_{HR_{Max}} - f_{HR_{Min}}) + f_{HR_{Min}}.
\end{align*}
In this model the variables have a directly interpretable mechanistic meaning: $SV$, $Pa$, $Pv$, $S$ which are respectively cardiac stroke volume (the amount of blood ejected by the heart), arterial blood pressure, venous blood pressure and autonomic barorelfex tone (the reflex responsible for adapting to perturbations in blood pressure, keeping homeostasis). \tabref{tab:cvs_details} in the appendix provides a full glossary of the above terms. The ODE state is $z_t = (SV(t), P_a(t), P_v(t), S(t))$. The observed state is the patient's vital signs and defined to be: $x_t = (P_a(t), P_v(t), f_{HR}(t))$, where $f_{HR}(t)$ is the patient's heart-rate at time $t$. Note that two of the three observed variables are the same as two of the latent variables, albeit with added noise as described in the dataset subsection. The CVS ODE model has several other parameters which we treat as known, setting them to the values stated in \citep{zenker2007inverse}.
The model is a 4-variable ODE system with 3-dimensional observations, which has much more involved dynamics than the pixel-pendulum above. 
Although far from comprehensive, this model can capture the prototypical behaviour of the cardiovascular system and its responses to pathological insults such as internal bleeding or septic shock that manifests as a reduction in peripheral vascular resistance. 
%We implemented here a slightly modified version of the original Zenker ODE. 
%\unote{we need to move the system to the appendix}
%using the following system:
%\begin{align*}
    % dSV/dt
%    \frac{dSV(t)}{dt} &= I_{external},  \\
    %
    % dPa/dt
%    \frac{dP_a(t)}{dt} &= \frac{1}{C_a}\left(\frac{P_a(t) - P_v(t)}{R_{TPR}(S)} - SV\cdot f_{HR}(S)\right) \\
    % dPv/dt
%    \frac{dP_v(t)}{dt} &= \frac{1}{C_v}\left(-C_a\frac{dP_a(t)}{dt} + I_{external} \right)\\
    %
    % dS/dt
%    \frac{dS(t)}{dt} &= \!\frac{1}{\tau_{Baro}}  \!
%    \left(    \! 1 \! - \! \frac{1}{1 + e^{-k_{width}(P_a(t) - P_{a_{set}})}   } \!-\!S\!  \right)\!,
%\end{align*}
%where 
%$R_{TPR}(S) = S(t)(R_{TPR_{Max}} - R_{TPR_{Min}}) 
 %   + R_{TPR_{Min}} + R_{TPR_{Mod}}$,  and  $ f_{HR}(S) = S(t)(f_{HR_{Max}} - f_{HR_{Min}}) + f_{HR_{Min}}$.

In this system we wish to find two parameters:\\ $\theta_{CVS} =  (I_{\text{external}}, R_{\text{Mod}})$ which are known-unknowns describing recognized clinical conditions: $I_{\text{external}}  <  0$ implies a patient is losing blood, and $R_{\text{Mod}}\geq 0$ implies that their total peripheral resistance is getting lower, which is a condition of distributive shock. Both conditions can lead to an observed drop in blood-pressure, and discerning the relative contribution of each to such a drop is important clinically: Often the underlying causes are not immediately clear and the choice of treatment relies on their accurate estimation. 

\textbf{Data Set}
We simulated 1000 sequences of length 400, with time steps of $\Delta_t = 1$. The parameter $I_{external}$ was randomly sampled to be either $-2$ or $0$, and the parameter $R_{TPR_{Mod}}$ was randomly sampled to be either $0.5$ or $0$. Initial ODE states uniformly sampled from $SV(0) \sim U[90, 100]$, $P_a(0) \sim U[75, 85]$, $P_v \sim [3, 7]$ and $S \sim [0.15, 0.25]$, where the intervals were set to the values given in \citet{zenker2007inverse}. The observations were additionally corrupted with white Gaussian noise with standard deviation of $\sigma_x = 5$ for $P_a$, $\sigma_x = 0.5$ for $P_v$ and $\sigma_x = 0.05$ for $f_{HR}$ (standard deviation matches scale of the observed signal). %An important note is that here the assumption that sparse latent observations are available is problematic: The reason is that measuring patients stroke-volume $SV(t)$ is considered a very hard task, and measuring $S(t)$ is impossible because it has no measurable meaning (it is a control signal).

% We simulated 1000 sequences of length 400. $I_{\text{external}}$ was randomly sampled to be either $-2$ or $0$, and $R_{\text{Mod}}$ was randomly sampled to be either $0.5$ or $0$. See appendix for complete details. 
% Initial ODE states uniformly sampled from $SV(0) \sim U[90, 100]$, $P_a(0) \sim U[75, 85]$, $P_v \sim [3, 7]$ and $S \sim [0.15, 0.25]$, where the intervals were set to the values given in \citep{zenker2007inverse}. Observations were additionally corrupted with white Gaussian noise with standard deviation $\sigma_x = 5$ for $P_a$, $\sigma_x = 0.5$ for $P_v$ and $\sigma_x = 0.05$ for $f_{HR}$ (standard deviation matches scale of the observed signal). %An important note is that here the assumption that sparse latent observations are available is problematic: The reason is that measuring patients stroke-volume $SV(t)$ is considered a very hard task, and measuring $S(t)$ is impossible because it has no measurable meaning (it is a control signal).

\textbf{Evaluation and results}
%In this scenario we cannot compare between our method and the DI baseline because it requires latent signal observations.
In addition to identification and extrapolation, we attempt to classify each $X^i$ series according to the sign of the inferred $I_{\text{external}}$ and $R_{\text{Mod}}$. These correspond to one of four possible clinical conditions: (1) Healthy (both non-negative), (2) Hemorrhagic shock ($I_{\text{external}}<0$, $R_{\text{Mod}}\geq 0$), (3) Distributive shock ($I_{\text{external}}\geq 0$, $R_{\text{Mod}}< 0$) and (4) Combined shock  ($I_{\text{external}}< 0$, $R_{\text{Mod}}< 0$). In addition to the previously mentioned baselines, we also compare  with the following baseline: cluster the observations using K-means with $K=4$, and assign each cluster to the most common true clinical condition. 

\tabref{tab:CVS.parameters} shows results on all the above tasks. L-ODE, and DI-1\% failed in this scenario and are not presented. We see that without any access to the latent space, GOKU-net successfully classifies which of the four clinical conditions the signal corresponds to, and extrapolates much better than the LSTM and DMM baselines. 
DI with mask rate lower than 5\% failed completely in reconstructing $X$.

%Figure \ref{fig:cvs_x_pred} shows how the extrapolation error of the observed signals evolves over time. 

\begin{table}
    \centering
    \begin{tabular}{ccccc} 
        \toprule
        Method & $I_{\text{external}}$ & $R_{\text{Mod}}$ & Class. & $X$ extrap.\\
        \midrule
        GOKU & 24 $\pm$ 2 & 3 $\pm$ .0 & 0\% & 29 $\pm$ 1\\
        K-Means & N/A & N/A & 13\%& N/A\\
        LSTM & N/A & N/A & N/A & 90 $\pm$ 2 \\
        DMM & N/A & N/A & N/A & 85 $\pm$ 3 \\
        DI-5\% & 1 $\pm$ 1 & 0 $\pm$ .0 & 0\% & 11 $\pm$ 10\\
    \bottomrule
    \end{tabular}
    \caption{CVS parameter identification and extrapolation error ($\times 10^{3}$), and classification error of clinical conditions. See text for K-means method. L-ODE and DI-1\% failed and are not shown.}
    \label{tab:CVS.parameters}
\end{table}

%%%%%%%%%%%%%%%%%%%%%%%%%%%%%%%%%%%

\section{Conclusion}
We explore the advantages of creating a model which is a hybrid of mechanistic and data-driven approaches \citep{baker2018mechanistic}. Compared to purely data-driven models such as Latent ODE \citep{chen2018neural} and LSTM, our model has an important advantage: it has access to part of the true mechanism underlying the data generating process, allowing us to reason about ``known-unknown'' variables which are often of crucial interest in clinical and other applications. 
%In our experiments we see the clear value of combining in the right way the known mechanistic model expressed in the ODE $f$, with the data-driven modeling of the emission function $g$. 
We show the extra knowledge encoded in the ODE structure allows us to correctly identify meaningful ODE parameters $\theta_f$, which is impossible with the methods above. We also show this extra knowledge translates to much better time-series extrapolation and significantly lower sample complexity. Comparing with direct inference of the ODE parameters, we show that adding the data-driven neural-net component allows GOKU-net to correctly learn the emission model, and yields much better estimates of the model parameters. %Finally, comparing to an augmented Latent ODE model with access to sparse observations of the true latent space, we still see a considerable advantage to our approach. 

Scientists have been developing mechanistic understanding of biological and physiological systems for centuries.
We believe finding new ways of combining this understanding into data-driven modeling is a valuable avenue for future research, especially in domains where data is scarce or where causal reasoning is crucial.

\section{Acknowledgments}
We wish to thank Guy Tennenholtz, Hagai Rossman and Rahul Krishnan for their useful comments on the manuscript. This research was partially supported by the
Israel Science Foundation (grant No. 1950/19).

\bibliographystyle{ACM-Reference-Format}
\bibliography{main}

\newpage
\onecolumn
\appendix
\section{Objective function}
\label{app.objective}
Derivation of the likelihood term in the objective (left term in Eq. \eqref{eq:obj}):
\begin{align*}
    \mathbb{E}_{q(Z, \tf, \ttf, \tz | X)} \left[\log p(X | Z, \tf, \tz, \ttf) \right]
    & = \mathbb{E}_{q(Z, \tf, \ttf, \tz | X)} \left[\log \prod_{t=0}^{T-1} p(x_t | z_t) \right] \\
    & = \sum_{t=0}^{T-1}\mathbb{E}_{q(Z, \tf, \ttf, \tz | X)} \left[\log p(x_t | z_t) \right] \\
    & = \sum_{t=0}^{T-1}\mathbb{E}_{q(z_t | X)} \left[\log p(x_t | z_t) \right].
\end{align*}
This follows since for $t' \neq t$: $x_t \independent (z_{t'}, \tf, \ttf, \tz) | z_t$.

Before decomposing the KL term, we note that the conditionals $p(z_0|\tz)$ and $p(\tf|\ttf)$ are deterministic, meaning they are Dirac functions with the peak defined by Eqs. \eqref{eq.hz} and \eqref{eq.htf}. The transition distribution $p(z_t|z_{t-1}, \tf)$ is also a Dirac function with the peak defined by Eq. \ref{eq.ode} as stated in Section \ref{sec:meth}.

Therefore, The KL term from Eq. \eqref{eq:obj} can be written as: 
\begin{align*}
    & KL \left[ q(Z, \tf, \tz, \ttf | X) ||  p(Z, \tf, \tz, \ttf)     \right] = \\
    & \int_{Z} \int_{\tf} \int_{\tz} \int_{\ttf} 
    q(\tz | X)    q(\ttf | X)    q(z_0 | \tz)    q(\tf | \ttf) 
    \prod_{t=1}^{T-1} q(z_t | z_{t-1}, \tf) \cdot \\
    & \cdot\log{\left[   \frac{p(\tz)    p(\ttf)    p(z_0 | \tz)    p(\tf | \ttf) 
    \prod_{t=1}^{T-1} p(z_t | z_{t-1}, \tf)}{q(\tz | X)    q(\ttf | X)    q(z_0 | \tz)    q(\tf | \ttf) 
    \prod_{t=1}^{T-1} q(z_t | z_{t-1}, \tf)}    \right]}.
\end{align*}
The KL term we got, decomposes into the sum of 3 terms:
\renewcommand{\theenumi}{(\roman{enumi})}%
\begin{enumerate}%[label=(\roman*),noitemsep,topsep=0pt]
\item The first term:
\begin{align*}
    & \int_{Z} \int_{\tf} \int_{\tz} \int_{\ttf} 
    q(\tz | X)    q(\ttf | X)    q(z_0 | \tz)    q(\tf | \ttf) 
    \prod_{t=1}^{T-1} q(z_t | z_{t-1}, \tf)
    \log{\left[   
    \frac{p(\tz)  \cancel{p(z_0 | \tz)}   }
    {q(\tz | X)  \cancel{q(z_0 | \tz)}   }
    \right]} = \\
    &   \int_{\tz} 
     q(\tz | X)  
    \log{\left[   \frac{p(\tz)}{q(\tz | X)}      \right]
    \int_{Z}  \int_{\tf} \int_{\ttf} q(\ttf | X)     q(z_0 | \tz)    q(\tf | \ttf) 
    \prod_{t=1}^{T-1} q(z_t | z_{t-1}, \tf)
    } =\\
    & KL \left( q(\tz | X) || p(\tz) \right),
\end{align*}
where $p(z_0|\tz)=q(z_0|\tz)$ by construction since both are determined exactly by Eq. \eqref{eq.hz}.
\item In the same way, we get:
\begin{align*}
    & \int_{Z} \int_{\tf} \int_{\tz} \int_{\ttf} 
    q(\tz | X)    q(\ttf | X)    q(z_0 | \tz)    q(\tf | \ttf) 
    \prod_{t=1}^{T-1} q(z_t | z_{t-1}, \tf)
    \log{\left[   
    \frac{p(\ttf)  p(\tf | \ttf)   }
    {q(\ttf | X)  q(\tf | \ttf)   }
    \right]} =\\
    & KL \left( q(\ttf | X) || p(\ttf) \right),
\end{align*}
where $p(\tf|\ttf)=q(\tf|\ttf)$ by construction since both are determined exactly by Eq. \eqref{eq.htf}.
\item The last term:
\begin{align*}
    & \int_{Z} \int_{\tf} \int_{\tz} \int_{\ttf} 
    q(\tz | X)    q(\ttf | X)    q(z_0 | \tz)    q(\tf | \ttf) 
    \prod_{t=1}^{T-1} q(z_t | z_{t-1}, \tf)
    \log{\left[   
    \frac{\prod_{t=1}^{T-1} p(z_t | z_{t-1}, \tf)}
    {\prod_{t=1}^{T-1} q(z_t | z_{t-1}, \tf)  }
    \right]} = 0,\\
\end{align*}
where $p(z_t | z_{t-1}, \tf) = q(z_t | z_{t-1}, \tf)$ by construction since both are determined exactly by the ODE system $f$. Thus the logarithmic term equals $0$.
\end{enumerate}

\section{Algorithms}
The algorithms below give the training procedure for a single iteration and a batch of size 1. The extension to larger batch sizes is straightforward. $X^i$ denotes a time sequence of length $T$ for an observed signal.

At inference time, we are given an observed signal $X^i$ of length $T$, and extrapolate the signal to time $T+\tau$. Meaning, the ODE-solver for-loop is from time $t=1$ to $t=T+\tau-1$, and the output sequence $\hat{X}^i$ is of length $T+\tau$.

\algref{alg:goku} gives the training procedure for GOKU-net. 

\algref{alg:di} gives the training procedure for the Direct Inference (DI) baseline described in Section \ref{sec:exp}. An important note is that this baseline is also given a latent space trajectory denoted as $Z^i$, and a grounding mask indicator denoted as $M^i$ (Eq. \eqref{eq.ground_loss}). These additional signals are not accessed by other methods. In this baseline, we first learn the parameters $\htf$ and the initial state of the ODE $\hat{z}_0$ of every signal in the train and test sets. We then evaluate $\hat{Z}^i$ for all signals in the train set, and use these predictions to learn the emission function $\hat{g}$, using the given train set signals $X^i$. In some cases, learning $\htz$ was too difficult for the baseline so we tried a different approach. We used the learned parameters $\htf$, the given ODE function $f$ and the first observed latent vector $z_{t'}^i$ (meaning the mask $M^i(t') = 1$ and $M^i(t<t') = 0$), and used the ODE solver to calculate $\htz$ backwards in time.

\algref{alg:goku-uu} addresses the unknown-unknowns task. This algorithm is very similar to \algref{alg:goku}, with the changes highlighted in blue. The main changes are that this method also includes an abstract function $f_{abs}$ which models the \emph{Unknown Unknowns} part of the ODE. I.e., the ODE is changed to be: $\frac{dz_t}{dt} = f_{ODE}(z_t, \tf) + f_{abs}(z_t)$.

\begin{algorithm}[h!]
\small
\caption{GOKU-net}
\label{alg:goku}
\begin{algorithmic}
    \STATE \textbf{Input:} 
    \STATE \qquad 1. sequence $X^i=(x_0,...,x_{T-1})$ 
    \STATE \qquad 2. ODE function $f$
    \STATE \qquad 3. ODE solver 
    \STATE \qquad 4. hyper-parameter $\lambda_1$
    \STATE initialize the neural nets $\phi^{enc}_{\tz}$, $\phi^{enc}_{\ttf}$, $h_z$, $h_{\theta}$ and $\hat{g}$.
    \STATE $[\mu_{\tz}, \sigma_{\tz}] = \phi^{enc}_{\tz}(X^i), \qquad \tz \sim \mathcal{N}(\mu_{\tz} , \sigma_{\tz}),
    \qquad \hat{z}_0 = h_z(\tz)$
    \STATE $[\mu_{\ttf}, \sigma_{\ttf}] = \phi^{enc}_{\ttf}(X^i), \qquad
    \ttf \sim \mathcal{N}(\mu_{\ttf}, \sigma_{\ttf}), \qquad
    \hat{\theta}_f = h_{\theta}(\ttf)
    $
    \FOR{$t=1,...,T-1$}
        \STATE $\hat{z}_t = ODE solver(f, \hat{\theta}_f, \hat{z}_{t-1})$
    \ENDFOR
    \STATE $\hat{X}^i = \hat{g}(\hat{Z}^i)$; \qquad \qquad 
    \algorithmiccomment{
    $\hat{Z}^i = (\hat{z}_0,...,\hat{z}_{T-1})$ for sample $i$}
    \STATE \emph{ll\_loss} = $\mathcal{L}_{likelihood}(X^i, \hat{X}^i)$;
    \qquad \algorithmiccomment{see first term in Eq. \eqref{eq:obj}}
    \STATE \emph{kl\_loss} = $\mathcal{L}_{kl}(\mu_{\tz}, \sigma_{\tz}, \mu_{\ttf}, \sigma_{\ttf})$;
    \qquad \algorithmiccomment{see second term in Eq. \eqref{eq:obj}}
    \STATE \emph{loss} = \emph{ll\_loss} + $\lambda_1$ \emph{kl\_loss}
    \STATE backpropagate(\emph{loss}) 
\end{algorithmic}
\end{algorithm}

\begin{algorithm}[h]
\small
\caption{Direct Inference (DI)}
\label{alg:di}
\begin{algorithmic}
    \STATE \textbf{Input:}
    \STATE \qquad 1. ODE function $f$
    \STATE \qquad 2. ODE solver
    \STATE \qquad 3. train and test sets of observed signals $X^i$, $Z^i$ and $M^i$.\\
    \FOR{train and test sets}
        \STATE Initialize
        $\hat{\theta}_f, \hat{z}_0$
        \FOR{$t=1,...,T-1$}
            \STATE $\hat{z}_t = ODE solver(f, \hat{\theta}_f, \hat{z}_{t-1})$
        \ENDFOR
        \STATE \emph{loss} =
        $||Z^i - \hat{Z}^i||_2$
        \STATE $\hat{\theta}_f := \hat{\theta}_f + \lambda \frac{\partial loss}{\partial \hat{\theta}_f}$ \qquad \algorithmiccomment{backpropagate loss through the ODE solver}
        \STATE $\hat{z}_0 := \hat{z}_0 + \lambda \frac{\partial loss}{\partial \hat{z}_0}$
    \ENDFOR
    
    \FOR{train set}
        \STATE $\hat{X}^i = \hat{g}(\hat{Z}^i)$
        \STATE \emph{generative\_loss} = $||X - \hat{X}^i||_2$
        \STATE backpropogate(\emph{generative\_loss})
    \ENDFOR
\end{algorithmic}
\end{algorithm}

\begin{algorithm}[h]
\small
\caption{GOKU with Unknown Unknowns (GOKU-UU)}
\label{alg:goku-uu}
\begin{algorithmic}
    \STATE \textbf{Input:} \\
    \qquad 1. sequence $X^i=(x_0,...,x_{T-1})$ \\
    \qquad 2. ODE function $f$\\
    \qquad 3. ODE solver\\
    \qquad 4. hyper-parameter $\lambda_1$

    \STATE initialize the neural nets
    {${\color{blue} f_{abs}}$},
    $\phi^{enc}_{\tz}$, $\phi^{enc}_{\ttf}$, $h_z$, $h_{\theta}$ and $\hat{g}$.
    \STATE $[\mu_{\tz}, \sigma_{\tz}] = \phi^{enc}_{\tz}(X^i), \qquad \tz \sim \mathcal{N}(\mu_{\tz} , \sigma_{\tz}),
    \qquad z_0 = h_z^{ODE}(\tz)$
    \STATE $[\mu_{\ttf}, \sigma_{\ttf}] = \phi^{enc}_{\ttf}(X^i), \qquad
    \ttf \sim \mathcal{N}(\mu_{\ttf}, \sigma_{\ttf}), \qquad
    \hat{\theta}_f = h^{ODE}_{\theta}(\ttf)$
    \FOR{$t=1,...,T-1$} 
        \STATE $z^{ODE}_t = ODE solver(f + $
        {\color{blue} $f_{abs}$} 
        $,\hat{\theta}_f, z_{t-1})$
    \ENDFOR
    \STATE $\hat{X}^i = \hat{g}(\hat{Z}^i)$
    \STATE \emph{loss} = $\mathcal{L}_{likelihood}(X^i, \hat{X}^i) + \lambda_1\mathcal{L}_{kl}(\mu_{\tz}, \sigma_{\tz}, \mu_{\ttf}, \sigma_{\ttf})$
    \qquad\algorithmiccomment{see Eq. \eqref{eq:obj}}
    % \STATE backpropagate(\emph{likelihood\_loss}, \emph{kl\_loss}, \emph{grounding\_loss}) 
    \STATE backpropagate(\emph{loss}) %w.r.t. neural nets ${\color{blue}f_{abs}}, \phi^{enc}_{\tz}$, $\phi^{enc}_{\ttf}$, $h_z$, $h_{\theta}, \hat{g}$
\end{algorithmic}
\end{algorithm}

\section{Experiments}
We provide here more information about the experiments described in Section \ref{sec:exp}. We ran all of the experiments on a desktop CPU. Full code implementation for creating the datasets, implementing GOKU-net, and implementing baselines is available on
{\urlstyle{sf} \textcolor{blue}{\url{github.com/orilinial/GOKU}}}.

\subsection{Hyper parameter selection}
All hyper parameters used for dataset creation, and for all methods, are explicitly declared in the code. In this section we describe the methods for selecting the hyper parameters.

\paragraph{KL distance}
In GOKU-net and L-ODE, we set the KL hyper parameter initial value to $10^{-5}$, with a KL annealing scheme so that it would gradually increase to the value of 1. To choose the KL hyper parameter initial value, we first set the KL hyper parameter to initial value of 1, and divided by 10 until finding the initial value that provided the best results on the validation set.

\paragraph{Optimization}
In all methods, we experimented with batch sizes from the set $\{16, 32, 64, 128\}$. We found that batch size of 64 provided the best results on the validation set for all methods in all experiments.
In all methods we used Adam optimizer with learning rate $10^{-3}$. We also experimented with other learning rates $10^{-2}$ and $10^{-4}$, which did not provide better results.

\subsection{Single Pendulum From Pixels}

\paragraph{Algorithm implementation details}
For all algorithms we used an input-to-RNN network and emission function exactly as suggested in Greydanus et al. \citep{greydanus2019hamiltonian}, composed of four fully-connected layers with ReLU activations and residual connections. The output of the input-to-RNN net dimension is 32.

In GOKU-net, for $\phi^{enc}_{\tz}$ we used an RNN with hidden dimension of 16 followed by a linear transformation to $\mu_{\tz}$ and another linear transformation to $\sigma_{\tz}$, both with dimension of 16. $\phi^{enc}_{\ttf}$ is very similar to $\phi^{enc}_{\tz}$ except for using a bi-directional LSTM instead of an RNN.
The $h$ functions are implemented with an MLP with one hidden layer with 200 neurons. The output of $h_{z}$ is linear, and the output of $h_{\theta}$ is forced to be positive using the softplus activation, so that $\theta_{pendulum}$ would be physically feasible. 
In Latent ODE, we used input dimension of 16 for the RNN, which then transforms linearly to $\mu_{z_0}$ and $\sigma_{z_0}$ with dimension of $16$ as well. The ODE function $f_{abs}$ is modeled as a neural network of sizes $4 \to 200 \to 200 \to 4$ with ReLU activation.
In the LSTM baseline, we used an LSTM with 4 layers and a hidden size of 16, followed by the same emission function as GOKU. In DMM, the inference model is built by first using the same input-to-RNN network as GOKU, followed by a combiner function that averages the previous $z_{t-1}$ and the RNN output $h_t$ to produce $z_{t}$. We set $z$ dim to be 32. The generative model is build by first using a transition function which is a 32->200->200->32 NN with ReLU activation function, and then using the same emission function as GOKU.
In HNN, We used the code provided by \citep{greydanus2019hamiltonian}. The only change we made is in the dataset creation process, $l$ is uniformly sampled instead of being constant.

\subsection{Pixel Pendulum with Unknown Unknowns}
In this experiment we aimed to show how GOKU can be modified to handle unknown unknowns in the ODE: We are given an ODE system that only partially describes the system that created the data. Specifically in this scenario, the pixel-pendulum data is created with a friction model:
\begin{align*}
    \frac{d\theta(t)}{dt} = \omega(t), \quad
    \frac{d\omega(t)}{dt} = -\frac{g}{l} \sin{\theta(t)} -\frac{b}{m}\omega(t),
\end{align*}
and we are only given with the friction-less ODE system in Eq. \eqref{eq.pendulum}. Our method (\algref{alg:goku-uu}) models the time derivatives of the unknown part, making the ODE functional form as:
\begin{equation*}
    \frac{dz_t}{dt} = f_{ODE}(z_t, \tf) + f_{abs}(z_t),
\end{equation*}
where $f_{abs}$ is modeled as a neural network.

\paragraph{Data set} We created this data set in the same way as in the friction-less pixel-pendulum experiment. Here we set $l \sim U[1, 2]$ as in the non-friction experiment, and we set in addition $m=1$, $b=0.7$.

\paragraph{Algorithm implementation details}
The only difference between this experiment and the pixel pendulum experiment, is that in GOKU-UU, we added a neural network that models $f_{abs}$, which is implemented as a fully connected network with $2 \to 200 \to 200 \to 2$ layers and ReLU activations. 

\begin{figure}[h]
    \centering
    \includegraphics[width=0.6\columnwidth]{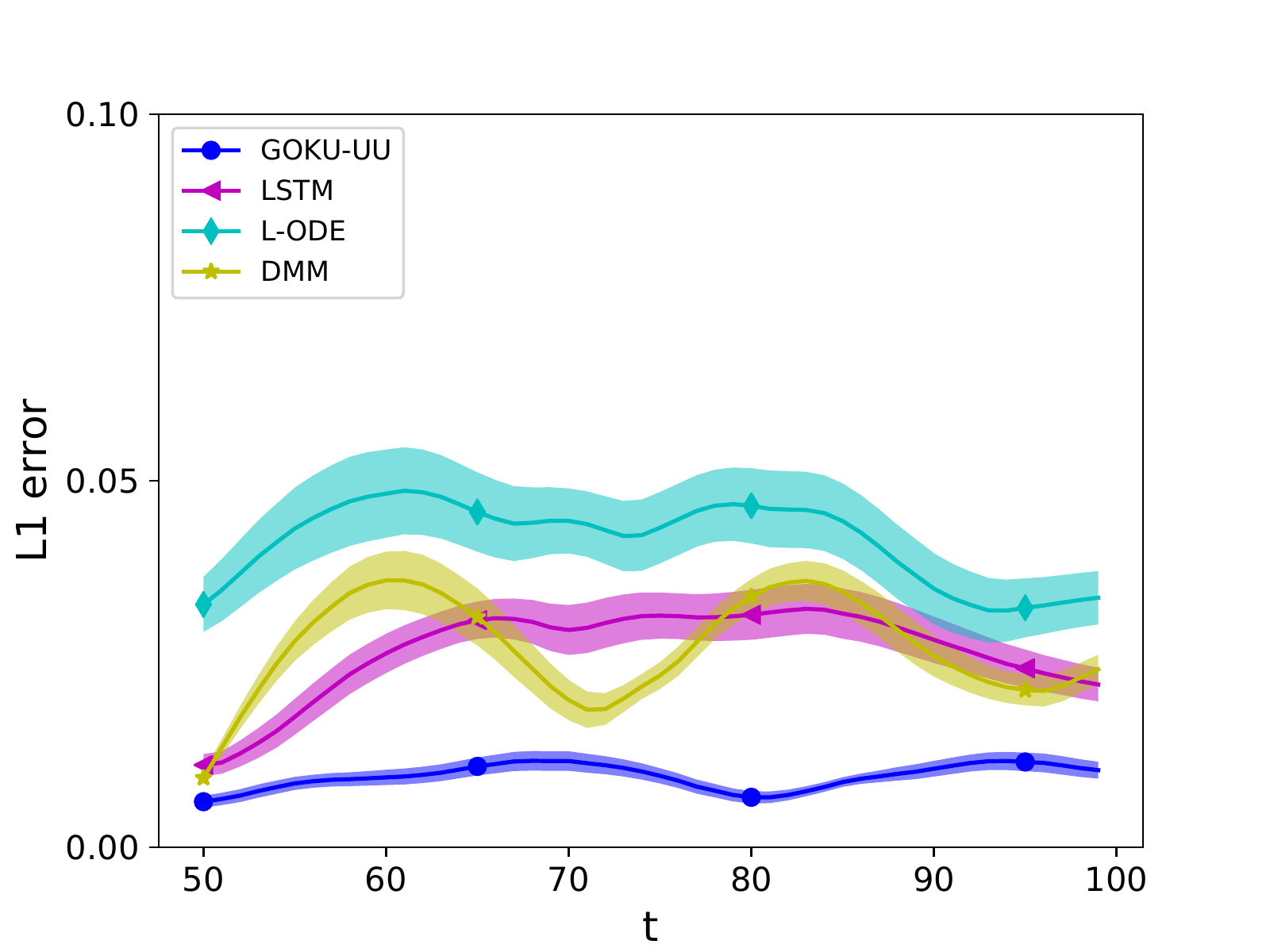}
    \caption{Pixel pendulum with friction - mean extrapolation error for observations $X$ over time steps after end of input sequence.}
    \label{fig:pp_friction_x}
\end{figure}

\begin{figure}[h]
    \centering
    \includegraphics[width=0.6\columnwidth]{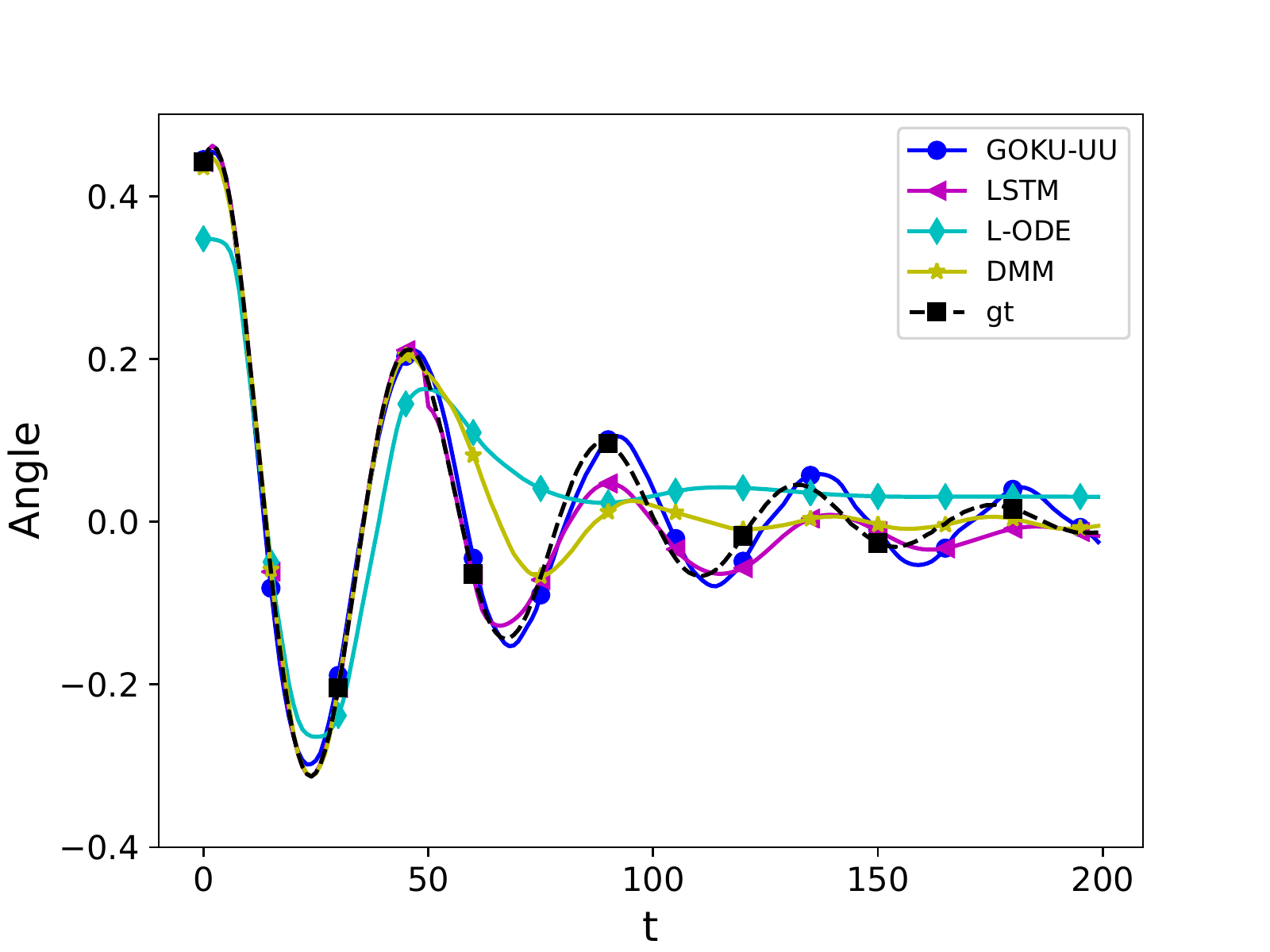}
    \caption{Pixel pendulum with friction predicted angle example. Comparing GOKU-UU to the baselines.}
    \label{fig:pp_friction_angle}
\end{figure}

\paragraph{Results}
In \figref{fig:pp_friction_x} we compare the $X$ extrapolation error between GOKU-UU and the baselines The observed signal is of length $T=50$, and we extrapolated the signals until $T=200$. In \figref{fig:pp_friction_angle} we demonstrate the extrapolation of $X$, by randomly selecting one test sample and showing the pendulum's predicted angle for future times. In both we observe that GOKU-UU achieved much better results than the compared baselines. In \figref{fig:pp_friction_angle_zeroed} we demonstrate that GOKU-UU's added function $f_{abs}$, learned only the friction part. This is done by first training using GOKU-UU, and then zeroing $f_{abs}$ during test time. \figref{fig:pp_friction_angle_zeroed} shows that the signal with $f_{abs}$ zeroed extrapolates as if there was no friction at all, suggesting that we successfully separated the friction model from the pendulum model.
We also tested if GOKU-UU could perform $\tf$ identification, and obtained the following results: correlation coefficient $=0.967$, and $L_1$ error of $0.109 \pm 0.013$.

These results show that using GOKU-net with the unknown-unknowns modification can successfully identify the ODE parameters and extrapolate the observed signal, although it does not observe the full ODE functional form. Moreover, it demonstrates capability to separate the Known-Unknowns (the given ODE's state and parameters) from the Unknown-Unknowns (the friction).

\subsection{Double Pendulum From Pixels}
In this experiment we used a double-pendulum ODE system, which has significantly more complicated dynamics than the the single pendulum one. Indeed, double pendulums are known to be chaotic systems. The ODE we used is exactly the one described in \verb|Acrobot-v1| environment from OpenAI Gym \citep{brockman2016openai}.
The ODE state in time $t$ is defined by the angles and angular velocities of the two pendulums: $z_t = (\theta_1, \omega_1, \theta_2, \omega_2)$. The ODE system is therefore:
\begin{align*}
    & \frac{d \theta_1 (t)}{dt} = \omega_1\\
    & \frac{d \theta_2 (t)}{dt} = \omega_2\\
    & \frac{d \omega_1 (t)}{dt} = -\left( d_2 \frac{d \omega_2 (t)}{dt} + \phi_1 \right) \frac{1}{d_1}\\
    &\frac{d \omega_2 (t)}{dt} = \frac{\frac{d_2}{d_1} \phi_1 - \phi_2}{m_2 c_2^2 + I_2 - \frac{d_2^2}{d_1}},
\end{align*}
where the auxilary functions $d_1, d_2, \phi_1, \phi_2$ are declared as:
\begin{align*}
    &d_1 = m_{1}c_{1}^2 + m_2(l_1^2 + (c_2)^2 + 2l_1 c_2  cos(\theta_2)) + I_1 + I_2 \\
    &d_2 = m_2(c_2^2 + l_{1}c_{2}cos(\theta_2)) + I_2 \\
    &\phi_2 = m_{2}c_{2} g cos(\theta_1 + \theta_2 - 0.5\pi)\\
    &\phi_1 = - m_{2}l_{1}c_{2} \omega_2^2 sin(\theta_2)
           - 2m_2 l_1 c_2 \omega_2 \omega_1 sin(\theta_2)
        + (m_1 l_1 + m_2 l_1)g cos(\theta_1 - 0.5\pi) + \phi_2.
\end{align*}
The parameters $m_1, m_2$ are the masses of the pendulums, $l_1, l_2$ are the lengths, $I_1, I_2$ are the moments of inertia, and $c_1, c_2$ are the position of the center of mass for each pendulum.

\paragraph{Dataset} We created this data set in a similar way to the pixel-pendulum experiment. For training we simulated 500 sequences of 50 time points, with time steps of $\Delta_t = 0.05$ and pre-processed the observed signals in exactly the same way as in the single pendulum experiment, but with frame size of $32 \times 32$. The ODE parameter we aimed to infer is $m_2$ and was uniformly sampled $m_2 \sim U[1,2]$. The other parameters were assumed to be constant and known as described in \verb|Acrobot-V1| (all parameters with values $1.0$, except $c_1 = c_2 = 0.5$ and $g = 9.8$). The ODE initial state $z_0$ was uniformly sampled, $z_0 \sim U[\frac{\pi}{10}, \frac{\pi}{6}]^4$. Each test set sequence is 100 time steps long, where the first 50 time steps are given as input, and the following 50 were used only for evaluating the signals extrapolation. 

\paragraph{Algorithm implementation details and Results} The implementation of this task is exactly as in the pixel pendulum task, except the obvious difference of incorporating the double pendulum ODE instead of the single pendulum one for GOKU and DI. Results are described fully in Section \ref{sec:exp}.

\subsection{CVS} 
We first give the complete ODE we used in this experiment. This is simplified form of the model given in Zenker et al. \citep{zenker2007inverse}:

\begin{align*}
    % dSV/dt
    \frac{dSV(t)}{dt} &= I_{\text{external}},  \\
    %
    % dPa/dt
    \frac{dP_a(t)}{dt} &= \frac{1}{C_a}\left(\frac{P_a(t) - P_v(t)}{R_{TPR}(S)} - SV\cdot f_{HR}(S)\right) \\
    % dPv/dt
    \frac{dP_v(t)}{dt} &= \frac{1}{C_v}\left(-C_a\frac{dP_a(t)}{dt} + I_{\text{external}} \right)\\
    %
    % dS/dt
   \frac{dS(t)}{dt} &= \!\frac{1}{\tau_{Baro}}  \!
    \left(    \! 1 \! - \! \frac{1}{1 + e^{-k_{width}(P_a(t) - P_{a_{set}})}   } \!-\!S\!  \right)\!,
\end{align*}
where 
\begin{align*}
    &R_{TPR}(S) = S(t)(R_{TPR_{Max}} - R_{TPR_{Min}}) 
    + R_{TPR_{Min}} + R_{TPR_{Mod}}\\
    &f_{HR}(S) = S(t)(f_{HR_{Max}} - f_{HR_{Min}}) + f_{HR_{Min}}.
\end{align*}

In this model the variables have a directly interpretable mechanistic meaning:  $SV$, $Pa$, $Pv$, $S$  are respectively cardiac stroke volume (the amount of blood ejected by the heart), arterial blood pressure, venous blood pressure and autonomic barorelfex tone (the reflex responsible for adapting to perturbations in blood pressure, keeping homeostasis). 
In \tabref{tab:cvs_details} we give information about the meaning of the ODE state variables and ODE parameters.

Beyond $I_{\text{external}}$ and $R_{TPR_{Mod}}$ (denoted $R_{\text{Mod}}$ for brevity in the main text),
the CVS ODE model has several other parameters which we treat as known, setting them to the values stated in \citet{zenker2007inverse}. The ODE state is $z_t = (SV(t), P_a(t), P_v(t), S(t))$. The observed state is the patient's vital signs and defined to be: $x_t = (P_a(t), P_v(t), f_{HR}(t))$, where $f_{HR}(t)$ is the patient's heart-rate at time $t$. Note that two of the three observed variables are the same as two of the latent variables, albeit with added noise as we explain now.

\paragraph{Dataset}
We simulated 1000 sequences of length 400. The parameter $I_{\text{external}}$ was randomly sampled to be either $-2$ or $0$, and the parameter $R_{TPR_{Mod}}$ was randomly sampled to be either $0.5$ or $0$. Initial ODE states uniformly sampled from $SV(0) \sim U[90, 100]$, $P_a(0) \sim U[75, 85]$, $P_v \sim [3, 7]$ and $S \sim [0.15, 0.25]$, where the intervals were set to the values given in \citep{zenker2007inverse}. Observations were additionally corrupted with white Gaussian noise with standard deviation $\sigma_x = 5$ for $P_a$, $\sigma_x = 0.5$ for $P_v$ and $\sigma_x = 0.05$ for $f_{HR}$ (standard deviation matches scale of the observed signal).

\paragraph{Algorithm implementation details}
For all algorithms we used an input-to-rnn network of 2 fully connected layers with ReLU activation and 64 hidden units, and output with dimension of 64.

In GOKU, The RNN and LSTM are implemented as in the LV experiment with output of dimension 64, followed by a linear transformation to $\mu_{z_0}$ and $\sigma_{z_0}$ of dimension 64 as well. The $h$ functions are implemented as in the LV, except that their output has a sigmoid activation layer, to bound them to a physically feasible solution. The emission function is a takes $P_a$ and $P_v$ from the latent trajectories, and a fully connected $4 \to 200 \to 1$ network with ReLU activation layer to compute $f_{HR}$.
In Latent ODE the emission function is a fully connected $4 \to 200 \to 3$ network with ReLU activation layer. In LSTM, we used the same network as in the LV experiment.
In DMM, the inference model is built by first using the same input-to-RNN network as GOKU, followed by a combiner function that averages the previous $z_{t-1}$ and the RNN output $h_t$ to produce $z_{t}$. We set $z$ dim to be 8. The generative model is build by first using a transition function which is a 8->200->200->8 NN with ReLU activation function, and then using the same emission function as GOKU.

\begin{table}[h]
    \centering
    \begin{tabular}{lL{10cm}l} 
        \toprule
        Symbol & Description & Unit \\
        \midrule
        $SV$ & Stroke volume, the volume of blood ejected during 1 cardiac cycle/ejection period & ml \\
        $P_a$ & Pressure in arterial compartment & mm Hg \\
        $P_v$ & Pressure in venous compartment & mm Hg \\
        $S$ & Autonomic barorelfex tone, i.e., the reflex responsible for adapting to perturbations in blood pressure, keeping homeostasis & -\\
        $f_{HR}$ & Heart rate, i.e., the number of complete cardiac cycles per unit time & Hz \\
        $R_{TPR}$ & Total peripheral/systemic vascular hydraulic resistance, i.e., the hydraulic resistance opposing the flow through the capillary streambed that is driven by the arterio–venous pressure difference & mm HG s/ml \\
        $C_a$, $C_v$ & Compliance of arterial, venous compartment &  ml/mm Hg \\
        $\tau_{Baro}$ & Time constant of the baroreflex response, i.e., of the linear low pass characteristic
        of the physiological negative feedback loop controlling arterial pressure & s \\
        $k_{width}$ & Constant determining the shape and maximal slope of the logistic baroreflex
        nonlinearity & mm Hg$^{-1}$\\
        $P_{a_{set}}$ & Set point of the baroreflex feedback loop & mm Hg \\
        $I_{\text{external}}$ & Possible external blood withdrawal or
        fluid infusion to or from the venous compartment & ml/s\\
        $R_{{TPR}_{Mod}}$ & Possible modification in $R_{TPR}$ & mm Hg s/ml \\
        \bottomrule
    \end{tabular}
    \caption{Glossary of Variables and Parameters of the Cardiovascular Model, as shown in Zenker et al. \citep{zenker2007inverse}}.
    \label{tab:cvs_details}
\end{table}

\end{document}